\documentclass[10pt,journal,compsoc]{IEEEtran}
\usepackage{amsmath,amsfonts}
\usepackage{algorithmic}
\usepackage{algorithm}
\usepackage{array}
\usepackage{textcomp}
\usepackage{stfloats}
\usepackage{url}
\usepackage{ragged2e}
\usepackage{verbatim}
\usepackage{graphicx}
\usepackage{subfigure}
\usepackage{booktabs}
\usepackage{float}
\usepackage{multirow}
\usepackage{cite}
\newtheorem{theorem}{Theorem}[section]
\newtheorem{corollary}{Corollary}[section]
\begin{document}

\title{Aggregation Weighting of Federated Learning via Generalization Bound Estimation}

\author{Mingwei~Xu,
	Xiaofeng~Cao$^*$, Ivor W.  Tsang, \IEEEmembership{Fellow IEEE}, and  James T. Kwok, \IEEEmembership{Fellow IEEE}
	
 \IEEEcompsocitemizethanks{\IEEEcompsocthanksitem \emph{Mingwei~Xu and Xiaofeng~Cao are with School of Artificial Intelligence, Jilin University, Changchun 130012, China. 
		E-mail: mingweixu20@outlook.com, xiaofengcao@jlu.edu.cn. Corresponding Author: Dr. Xiaofeng Cao}}\protect 
 \IEEEcompsocitemizethanks{\IEEEcompsocthanksitem \emph{I. W. Tsang is with  the Centre for Frontier AI Research (CFAR) and Institute of High Performance Computing,  Agency for Science, Technology and Research (A$^*$STAR), Singapore. 
E-mail: ivor.tsang@uts.edu.au.   }}\protect

\IEEEcompsocitemizethanks{\IEEEcompsocthanksitem \emph{J. T. Kwok is with the Department of Computer Science
and Engineering, The Hong Kong University of Science and Technology,
Hong Kong, China. Email: jamesk@cse.ust.hk.}}\protect

\IEEEcompsocitemizethanks{\IEEEcompsocthanksitem \emph{  This work was supported in part by National Natural Science Foundation of China (Grant Number: 62206108), in part by Maritime AI Research Programme (SMI-2022-MTP-06) and AI Singapore OTTC Grant (AISG2-TC-2022-006), and in part by the Research Grants Council of the Hong Kong Special Administrative Region (Grant 16200021). }}\protect 
}
\markboth{Journal of \LaTeX\ Class Files,~Vol.~14, No.~8, August~2015}%
{Shell \MakeLowercase{\textit{et al.}}: Bare Demo of IEEEtran.cls for Computer Society Journals}


\IEEEtitleabstractindextext{\begin{abstract} \justifying  
		Federated Learning (FL) typically aggregates client model parameters using a weighting approach determined by sample proportions. However, this naive weighting method may lead to unfairness and degradation in model performance due to statistical heterogeneity and the inclusion of noisy data among clients. Theoretically, distributional robustness analysis has shown that the generalization performance of a learning model with respect to any shifted distribution is bounded. This motivates us to reconsider the weighting approach in federated learning. In this paper, we replace the aforementioned weighting method with a new strategy that considers the generalization bounds of each local model. Specifically, we estimate the upper and lower bounds of the second-order origin moment of the shifted distribution for the current local model, and then use these bounds disagreements as the aggregation proportions for weightings in each communication round. Experiments demonstrate that the proposed weighting strategy significantly improves the performance of several representative FL algorithms on benchmark datasets. 
	\end{abstract}
	
	}
\maketitle
\IEEEdisplaynontitleabstractindextext

\section{Introduction}
\IEEEPARstart{D}{ata} security and privacy protection are key subjects in the fields of data mining and machine learning. Federated learning (FL) based on the distributed machine learning framework that trains models across multiple decentralized clients was thus proposed~\cite{mcmahan2017communication} and sparked a research boom in related research communities~\cite{kairouz2021advances,9238427,10274102,pmlr-v202-guo23b}. In this federated learning framework, it only requires parametric communications between local clients and global servers without data sharing, and has recently been applied in many areas, including the Internet of Things (IoT)~\cite{Li9741255,ji2022fedqnn}, computer vision~\cite{liu2020fedvision,Xu_2022_CVPR}, automatic driving~\cite{pokhrel2020federated,9827020}, medical health~\cite{kaissis2021end,xu2021federated}.

Statistical heterogeneity presents a significant challenge in federated learning as it indicates that the data on different clients follow non-independent and non-identical distributions (Non-IID). Heterogeneity in federated learning~\cite{li2021federated} typically arises when clients possess noisy and class-imbalanced data. As a result, the optimization directions of local and global empirical loss functions become inconsistent, leading to a considerable decline in model performance and hindering convergence to the optimal solution. This inconsistency often necessitates increased communication requirements in federated learning systems. To address the aforementioned issues stemming from statistical heterogeneity, numerous early explorations have been conducted in federated learning. These explorations primarily focus on mitigating model drift and constraining local updates to prevent excessive divergence from the global model. For example, FedProx\cite{li2020federated} introduces a regularization term in the loss function, while SCAFFOLD~\cite{pmlr-v119-karimireddy20a} incorporates global and local control variables to alleviate deviation. FedDyn~\cite{acar2021federated} provides a dynamic regularizer for each client from a different perspective. These FL heterogeneity research efforts rely on aggregating client model parameters on the server using a weighting mechanism determined by local sample proportions. However, the naive weighting approach may introduce unfairness and less robustness due to statistical heterogeneity across clients, highlighting the need to reconsider the weighting mechanism for federated learning.

Reweighting has proven to be an effective and robust training technique in the machine learning community, particularly when dealing with noisy and imbalanced training data~\cite{ren2018learning,shu2019meta}. In the context of noisy training~\cite{jiang2014self,wang2017robust}, clean training data is assigned larger weights based on their lower training loss, while noisy data is assigned lower weights due to its potential to perturb the training model. Similarly, in the case of imbalanced training~\cite{sun2007cost,malisiewicz2011ensemble,lin2017focal}, classes with fewer samples are assigned higher weights to compensate for the lack of training information and abnormal loss values. However, in certain scenarios where both noisy and imbalanced data coexist among local clients in federated learning, these reweighting strategies may not be sufficiently robust to handle the complexity of FL heterogeneous data. 

\textbf{Motivation.} In the analysis of distributional robustness, the generalization performance of a shifted and heterogeneous data distribution can be bounded, providing control over the worst-case performance of any model-agnostic training model. Specifically, this generalization bound is positively correlated with the degree of the shifted distribution. In other words, the more heterogeneous the distribution is, the more challenging it becomes to accurately estimate the generalization performance. Building upon these insights, we propose a novel weighting strategy for parameter aggregation in federated learning that leverages the bound disagreement of shifted heterogeneous distributions \cite{weber2022certifying}. The estimation of bound disagreements theoretically reflects the training difficulty within a client's data distribution. A tighter bound disagreement indicates a more robust training performance. In this setting, any client with a shifted distribution exhibiting significant heterogeneity should be assigned a smaller weighting due to the large generalization disagreement it has with the server. By considering the bound disagreement, our proposed weighting strategy seeks to improve the robustness and fairness of parameter aggregation in federated learning. 

\textbf{Insight.} Theoretically, both the first-order origin moment and the second-order origin moment are expectations of different forms of the robustness loss function, while the key difference is that the second-order origin moment is flatter than the first-order origin moment for loss values below 1 and exhibits stronger convexity for values above 1. In the context of  Sharpness-Aware Minimization(SAM)~\cite{foret2021sharpnessaware}, it is believed that a flat minimum is preferred over a sharp minimum, as flat minima tend to be more stable. In robust weighting of FL, we aim to converge to a flat minimum, in which the second-order origin moment aligns with this purpose. Therefore, in the process of estimating the generalization bound, we utilize the second-order origin moment for bound generalization.

\textbf{Contributions}. In this paper, instead of relying on sample proportion weighting, we introduce a weighting scheme based on the estimation of generalization bounds for local models. Specifically, we leverage the superior flatness and convexity of the aforementioned second-order origin moment to uniquely estimate the generalization bound. By computing bound disagreements, we dynamically adjust the aggregation weightings in each communication round, ensuring a fairer client participation in the training process. Clients with tighter bound disagreements are assigned higher aggregate weightings, indicating a higher level of homogeneity. The main contributions of this work can be summarized as follows:

\begin{enumerate}
	\item  \textbf{Perspective from distributional robustness.} We reconsider the aggregation weighting approach in federated learning from the perspective of distributional robustness, which enables the bounding of generalization performance for any shifted distribution of local models.
	
	\item \textbf{New theoretical insight.} Theoretically, we utilize the second-order origin moment of the loss function, which exhibits better generalization performance compared to the first-order origin moment and avoids the aggregation weightings approaching zero at sharp values. Specifically, we provide upper and lower bounds on the generalization performance metric under data distribution shift, where we employ the second-order moment, denoted as $\mathbf{E}_{Q}[l^2(\cdot)]$ ($l$ is the loss). From the perspective of the Bias-Variance trade-off analysis, this second-order moment approximates the sum of squared bias and variance, $\mathbf{E}_{Q}[l^2(\cdot)] = \mathbf{E}^2_{Q}[l(\cdot)] + \mathbf{V}[l (\cdot)]$.
  	
	\item  \textbf{Robust aggregation weighting.} We propose a novel approach to address the inherent unfairness in the traditional sample proportion weighting used in federated learning. Our strategy involves estimating generalization bounds and implementing bound disagreements weighting scheme that improves aggregation efficiency in the presence of statistical heterogeneity. We extensively evaluate our approach using popular federated learning algorithms, including FedAvg~\cite{mcmahan2017communication}, FedProx~\cite{li2020federated}, SCAFFOLD~\cite{pmlr-v119-karimireddy20a}, and FedDyn~\cite{acar2021federated}. The experimental results demonstrate significant improvements achieved through our proposed approach.
\end{enumerate}	

The organization of this paper is as follows. In Section 2, we introduce research work closely related to the issues addressed in this paper. The Section 3 covers the theoretical foundations of our work, including the Bias-Variance trade-off and Distributional Robustness Analysis. In Section 4, we provide the specific algorithmic process. We start by establishing a basic framework for federated learning, offering upper and lower bound estimates for generalization performance using second-order origin moment, along with relevant theorem proofs. Based on this, we construct bound disagreements as a visual representation of weighting aggregation. In Section 5, we validate the proposed algorithm on multiple baselines and datasets, providing specific experimental analyses. In Section 6, we summarize the work presented in this paper.

\section{Related Work}
Federated Learning is widely recognized for its ability to preserve data privacy by aggregating local models without sharing raw data. Research in the field primarily focuses on three key aspects: privacy and security~\cite{orekondy2018gradient,zhu2019deep,bagdasaryan2020backdoor}, communication efficiency~\cite{li2020communication,reisizadeh2020fedpaq,wu2022communication}, and heterogeneity~\cite{mohri2019agnostic,haddadpour2021federated,Ma2022ContinualFL}.

\textbf{Heterogeneity in Federated Learning.} Kairouz et al.\cite{kairouz2021advances} provided a classification of Non-IID scenarios, highlighting five different situations of Non-IID data distribution: (1) Feature distribution skew, (2) Label distribution skew, (3) Same label, different features, (4) Same features but different labels, and (5) Quantity skew or unbalancedness. Furthermore, Li et al.\cite{li2021federated} proposed a benchmark for partition strategies, offering comprehensive guidelines and datasets that cover various Non-IID scenarios. Zhao et al.\cite{zhao2018federated} introduced a solution involving the generation of a shared small dataset on the server, which is used for initial model training before being sent to the clients along with the published global model. Li et al.\cite{li2019convergence} conducted convergence analysis of FedAvg on Non-IID datasets, recognizing that the heterogeneity of the dataset slows down the convergence rate and may lead to deviation from the optimal solution. They proposed controlling the deviation by attenuating the learning rate. Luo et al.~\cite{luo2021no} extensively studied the implicit representations of different layers in neural networks and identified the presence of larger bias in the classifier as the main reason for degraded performance on Non-IID data. While previous works have addressed the challenges of statistical heterogeneity in federated learning from various perspectives, they commonly employ the typical sample proportion for aggregating local models without exploring other methods. In this paper, we revisit the weighting approach and specifically focus on robust aggregation weighting.

\textbf{Robust Reweighting.}  Robust reweighting is a widely used concept in machine learning. To enhance out-of-distribution (OOD) generalization and mitigate overfitting in large overparameterized models, Zhou et al.\cite{pmlr-v162-zhou22d} proposed an effective reweighting of training samples. By applying standard empirical risk minimization training on the weighted training data, superior OOD generalization performance can be achieved. Addressing the issue of collinearity among input variables, which can inflate parameter estimation errors, Shen et al.\cite{shen2020stable} developed a sample reweighting technique that reduces collinearity. Some recent works have employed unbiased validation sets for sample reweighting. In research by Ren et al.\cite{ren2018learning}, a meta-gradient descent step was performed on the current mini-batch sample weightings to minimize the loss on a clean and unbiased validation set, determining the sample weightings for the mini-batch. Another work by Shu et al.\cite{shu2019meta} introduced a meta-learning method to reweight the loss. By training an MLP network on a small unbiased validation set, where the loss value served as input and the corresponding loss value weighting was produced as output, the neural network learned how to reweight different data losses. In the context of federated Learning, Pillutla et al.\cite{pillutla2022robust} presented a novel method that utilized the geometric median for aggregating updates, increasing the robustness of the aggregation process against potential poisoning of local data or model parameters. Additionally, Li et al.\cite{li2022communication} focused on the detrimental effects of corrupted labels in federated learning and proposed a learning-based reweighting approach to mitigate the impact of noisy labels.

\section{Preliminary }

\subsection{Bias and Variance Trade-off}
In machine learning, the expected cost of a trained model is divided into three non-negative components: intrinsic target noise, squared bias, and variance \cite{kohavi1996bias}. The trade-off between bias and variance is a statistical tool that aids in understanding the generalization of a trained model's predictions \cite{yang2020rethinking}. An optimal trade-off leads to a more accurate model that avoids both overfitting and underfitting.

The training dataset $D = {(x_1, y_1), \dots, (x_n, y_n)}$ consists of independent and identically distributed samples drawn from a distribution $P(X, Y)$. Here, $\mathbf{x}$ represents a test sample, and $y$ denotes its true label. $h_{D}(\mathbf{x})$ represents a hypothesis learned by a machine learning algorithm on the dataset $D$, while $h(\mathbf{x})$ denotes the expected label given input $\mathbf{x}$. The expected model hypothesis is denoted as $\bar{h} = E_{D \sim P^n} \left[ h_D \right] = \frac{1}{N} \sum h_{D}(\mathbf{x})$, and the expected test error is represented by $E_{(\mathbf{x}, y) \sim P} \left[ \left(h_D (\mathbf{x}) - y \right)^2 \right]$.


The bias-variance decomposition as follows:
\begin{equation}
	\begin{aligned}
		& \underbrace{\mathbf{E}_{\mathbf{x}, y, D} \left[\left( y - h_{D}(\mathbf{x})  \right)^{2}\right]}_\mathrm{Expected\;Test\;Error} \\
		& =   \underbrace{\mathbf{E}_{\mathbf{x}, D}\left[\left(h_{D}(\mathbf{x}) - \bar{h}(\mathbf{x})\right)^{2}\right]}_\mathrm{Variance} \\
		& + \underbrace{\mathbf{E}_{\mathbf{x}} \left[\left(\bar{h}(\mathbf{x}) - h(\mathbf{x})  \right)^{2}\right]}_\mathrm{Bias^2}  + \underbrace{\mathbf{E}_{\mathbf{x}, y} \left[\left( h(\mathbf{x}) -y\right) ^{2}\right]}_\mathrm{Noise}	,		
	\end{aligned}
\end{equation}

Derivation of the bias-variance decomposition can be found in Appendix \ref{appendix:A}.

In practical applications, the noise term is usually difficult to detect. We usually set it as constant. Therefore, in this paper, we approximate the expected test error as:
\begin{equation}
	\begin{aligned}
		& \mathbf{E}_{\mathbf{x}, y, D} \left[\left( y - h_{D}(\mathbf{x})  \right)^{2}\right] \\
		& \approx    \mathbf{E}_{\mathbf{x}, D}\left[\left(h_{D}(\mathbf{x}) - \bar{h}(\mathbf{x})\right)^{2}\right]
		+  \mathbf{E}_{\mathbf{x}} \left[\left(\bar{h}(\mathbf{x}) - h(\mathbf{x})  \right)^{2}\right] ,
	\end{aligned}
\end{equation}
Then
\begin{equation}
	\begin{aligned}
		& \mathbf{E}_{\mathbf{x}, y, D} \left[\left( y - h_{D}(\mathbf{x})  \right)^{2}\right]  \\
		& \approx \mathbf{E}_{\mathbf{x},y, D}\left[\left((h_{D}(\mathbf{x}) - y) - (\bar{h}(\mathbf{x}) - y) \right)^{2}\right] \\
		& + \mathbf{E}_{\mathbf{x}} \left[\left(\bar{h}(\mathbf{x}) - h(\mathbf{x})  \right)^{2}\right],
	\end{aligned} \label{Exy}
\end{equation}
Let $l(Z) = h_{D}(\mathbf{x}) - y$, the above equation~(\ref{Exy}) can be written as:
\begin{equation}
	\begin{aligned}
		& \mathbf{E}_{Z, D} \left[l^{2}(Z) \right] \\
		&\approx  \mathbf{E}_{\mathbf{x}} \left[\left(\bar{h}(\mathbf{x}) - h(\mathbf{x})  \right)^{2}\right]  + \mathbf{E}_{Z, D}\left[\left( l(Z) - \bar{l}(Z) \right)^{2}\right] ,
	\end{aligned}
\end{equation}
According to the law of large numbers, when $N$ is sufficiently large, $\bar{h}(\mathbf{x}) - h(\mathbf{x}) \approx h_{D}(\mathbf{x}) - y$, and the bias is approximately equal to $\mathbf{E}\left[l(Z)\right]$. For the sake of simplicity, we can rewrite the expected test error as follows:
\begin{equation}
	\mathbf{E}\left[l^2(Z)\right] \approx \mathbf{E}^2 \left[l(Z)\right] + \mathbf{V} \left[l(Z)\right] \label{El2} .
\end{equation}
Equation (\ref{El2}) also satisfies the variance formula $V[\mathbf{x}] = E[\mathbf{x}^2] - E^2[\mathbf{x}]$ in statistics. However, it is worth noting that this equation incorporates both bias and variance. When utilizing $\mathbf{E}[l^2(Z)]$ as a representation of generalization performance to estimate bound disagreements, this provides us with an additional perspective of understanding.

\subsection{Distributional Robustness Analysis}

\textbf{Distributional Robustness.} Distributional robustness optimization~\cite{Duchi2021,gao2022distributionally} is a technique that enhances the robustness of a model by optimizing for the worst-case distribution. In this approach, we consider $x \in X$ as the input, $y \in Y$ as the output from the joint data distribution $\mathbf{P}(X,Y)$, and $\mathbf{h}: X \rightarrow Y$ as the machine learning model. Given a loss function $\mathcal{L}: Y \times Y \rightarrow \mathbf{R}_{+}$, the objective is to minimize the following expression:
\begin{equation}
	\inf _{\theta \in \mathbf{R}^n} \sup _{\mathbf{Q} \in \mathcal{U}_P} \mathbf{E}_{(X, Y) \sim \mathbf{Q}}[\mathcal{L}(\mathbf{h}_{\theta}(X), Y)],
\end{equation}
Here, $\mathcal{U}_P \subseteq \mathcal{P}(X,Y)$ represents a set of uncertainty probability distributions. By solving this optimization problem, we can obtain model parameters that provide sufficient robustness.

On the distributional robustness framework, Werber et al.~\cite{weber2022certifying} investigated the disagreement in generalization performance among agnostic models resulting from discrepancies in data distribution. They provided upper and lower bounds for model generalization performance. Drawing inspiration from Theorem 2.2 in Werber's work and our above analysis, we extend the use of the second-order origin moment instead of the first-order moment mentioned in the original text. By introducing distance parameter $\epsilon$, we establish a different way to bound the robust performance of $\mathbf{h}$ on a shifted data distribution $\mathbf{Q}$: 
\begin{equation}
	\begin{aligned}
		& \forall \mathbf{Q}: Hellinger\_Dist (\mathbf{Q}, \mathbf{P}) \leq \epsilon  \\ 
		& \Rightarrow \mathbf{E}_{(X, Y) \sim \mathbf{Q}}[\mathcal{L}^2(\mathbf{h}_{\theta}(X), Y)] \leq B_{\mathcal{L}^2}(\epsilon, \mathbf{P}),
	\end{aligned}
\end{equation}
Here, $\mathbf{P}$ denotes the actual distribution, and $B_\mathcal{L}^2$ represents a bound that depends on the distance $\epsilon$ and the current data distribution $\mathbf{P}$. The $Hellinger\_Dist(,)$ refers to the Hellinger distance, a measure used to quantify the similarity between two probability distributions in machine learning. 

The upper and lower bounds of $\mathbf{E}_{(X, Y) \sim \mathbf{Q}}[\mathcal{L}^2(\mathbf{h}(X), Y)]$  follow from \textbf{Theorem 3.1, Theorem 3.2}.
\begin{theorem}
	Upper bound of generalization performance for agnostic model under shifted distribution: Let $\mathcal{L}: Y \times Y \rightarrow \mathbf{R}_{+}$ be a loss function, for some $M>0$ assume that $\sup _{(x,y) \in (X,Y)}|\mathcal{L}(\mathbf{h}(x), y)| \leq M$ . Then $\sup _{(x,y) \in (X,Y)}|\mathcal{L}^2(\mathbf{h}(x), y)| \leq M^2$, for any probability measure $\mathbf{P}$ on (X,Y) and $\epsilon>0$, we have:
	\begin{equation} 
		\begin{aligned}
			& \sup_{\mathbf{Q} \in B_\rho(\mathbf{P})} \mathbf{E}_\mathbf{Q}[\mathcal{L}^2(\mathbf{h}(X), Y)]  \leq  \mathbf{E}_\mathbf{P}[\mathcal{L}^2(\mathbf{h}(X), Y)]  + \\
			& 2 \lambda_\epsilon \left[\mathbf{V}_\mathbf{P}[\mathcal{L}^2(\mathbf{h}(X), Y)]\right]^{1/2} +\Delta^{u}_{\epsilon} ,
		\end{aligned}
	\end{equation} 
	$$
	\begin{aligned}
		& \Delta^{u}_{\epsilon} = \epsilon^2\left(2-\epsilon^2\right)\left[ M^2-\mathbf{E}_\mathbf{P}[\mathcal{L}^2(\mathbf{h}(X), Y)]\right. \\ &\left.-\frac{\mathbf{V}_\mathbf{P}[\mathcal{L}^2(\mathbf{h}(X), Y)]}{M^2-\mathbf{E}_\mathbf{P}[\mathcal{L}^2(\mathbf{h}(X), Y)]}\right],	
	\end{aligned} 	
	$$
\end{theorem}
where $\lambda_\epsilon=\left[\epsilon^2\left(2-\epsilon^2\right)\left(1-\epsilon^2\right)^2\right]^{1/2}$ and $B_\epsilon(\mathbf{P})=\{\mathbf{Q} \in$$\mathbf{P}(X,Y): H(\mathbf{P}, \mathbf{Q}) \leq \epsilon\}$ is the Hellinger ball of radius $\epsilon$ centered at $\mathbf{P}$. The radius $\epsilon$ is required to be 
\begin{equation}
	\epsilon^2 \leq 1-\left[ 1+ \left(\frac{M^2-\mathbf{E}_\mathbf{P}[\mathcal{L}^2(\mathbf{h}(X), Y)]}{\sqrt{\mathbf{V}_\mathbf{P}[\mathcal{L}^2(\mathbf{h}(X), Y)]}} \right)^2 \right]^{-1 / 2}.
\end{equation}
\begin{theorem}
	Lower bound of generalization performance for agnostic model under shifted distribution: Let $\mathcal{L}: Y \times Y \rightarrow \mathbf{R}_{+}$ be a nonnegative function taking values in $(X,Y)$. Then, for any probability measure $\mathbf{P}$ on $(X,Y)$ and $\epsilon>0$, we have:
	\begin{equation}
		\begin{aligned}
			& \inf _{\mathbf{Q} \in B_\epsilon(\mathbf{P})} \mathbf{E}_\mathbf{Q}[\mathcal{L}^2(\mathbf{h}(X), Y)] \geq \mathbf{E}_\mathbf{P}[\mathcal{L}^2(\mathbf{h}(X), Y)] - \\
			& 2 \lambda_\epsilon \left[\mathbf{V}_\mathbf{P}[\mathcal{L}^2(\mathbf{h}(X), Y)]\right]^{1/2}-\Delta^{l}_{\epsilon} , 
		\end{aligned}		
	\end{equation}
	\centering{$\Delta^{l}_{\epsilon} = \epsilon^2\left(2-\epsilon^2\right)\left[\mathbf{E}_\mathbf{P}[\mathcal{L}^2(\mathbf{h}(X), Y)]-\frac{\mathbf{V}_\mathbf{P}[\mathcal{L}^2(\mathbf{h}(X), Y)]}{\mathbf{E}_\mathbf{P}[\mathcal{L}^2(\mathbf{h}(X), Y)]}\right]$},
\end{theorem}
where $\lambda_\epsilon=\left[\epsilon^2\left(1-\epsilon^2\right)^2\left(2-\epsilon^2\right)\right]^{1/2}$ and $B_\epsilon(\mathbf{P})=\{\mathbf{Q} \in \mathcal{P}(X,Y): H(\mathbf{P}, \mathbf{Q}) \leq \epsilon\}$ is the Hellinger ball of radius $\epsilon$ centered at $\mathbf{P}$. The radius $\epsilon$ is required to be small enough such that
\begin{equation}
	\epsilon^2 \leq 1-\left[1+ \left( \frac{\mathbf{E}_\mathbf{P}[\mathcal{L}^2(\mathbf{h}(X), Y)]}{\sqrt{\mathbf{V}_\mathbf{P}[\mathcal{L}^2(\mathbf{h}(X), Y)]}} \right) ^2\right]^{-1 / 2} .
\end{equation}
The proof of \textbf{Theorem 3.1} and \textbf{Theorem 3.2} is available in Appendix \ref{appendix:B}.

The aforementioned \textbf{Theorem 3.1} and \textbf{Theorem 3.2} provide upper and lower bounds for the generalization performance of the agnostic model in the presence of data distribution disagreement $\epsilon$. Its upper and lower bounds are a combination of expectations and variances. It can be regarded as the calculation of the second-order origin moment of the loss function with variance as a regularization term. Regarding the local model in Federated Learning, we can estimate its bounds on generalization performance disagreement by leveraging the upper and lower bounds mentioned earlier.

\section{Federated Learning with Robust Weighting}

\textbf{Problem Statement.} In typical Federated Learning studies, the weighting proportions assigned to local models during aggregation adhere to the principle: $\sum_{k=1}^K p_k=1$, where $p_k$ represents the proportion of local training samples relative to the total training samples. This approach ensures that each local model's contribution is appropriately considered. However, in heterogeneous scenarios, where data distributions may differ across local models, the strategy of determining aggregation weightings based on sample proportions takes into account the potential adverse effects caused by heterogeneous data.

The trade-off between bias and variance demonstrates that the second-order origin moment consist of important statistical indicators, namely bias and variance, which provide valuable insights into the accuracy and generalization capacity of a learning model. More importantly, according to the analysis of Sharpness-Aware Minimization mentioned earlier, the second-order origin moment demonstrates better stability and convexity. Based on the aforementioned analysis, our objective is to estimate the upper and lower bounds of the second-order origin moment of the local models. These bounds are obtained under a distributed robust setting. By doing so, we aim to gain a comprehensive understanding of the model's weighting aggregation performance and account for potential variations and uncertainties.

We present the formal problem statement concisely as follows: to mitigate the adverse effects of heterogeneous data in parameter aggregation weighting, we initially assign a pre-defined distance to quantify the disagreement in data distribution, representing the degree of distribution shift. Subsequently, we estimate the upper and lower bounds of generalization for the local models. Finally, the disagreements in the generalization bounds provide the foundation for determining the weightings used in aggregation.

\subsection{Federated Learning}

\textbf{General Federated Learning.} In typical FL~\cite{mcmahan2017communication}, the learned objective can be generalized as an optimization function:
\begin{equation}
	\min_{\mathbf{h}} L(\mathbf{h}) = \sum_{k=1}^K p_k L_k(\mathbf{h}),
\end{equation} 
where $L_k$ denotes the total training loss of the $k$-th clients, $\mathbf{h}$ denotes the hypothesis of learning model, and $K$ denotes the number of local clients participating in the training. Assume that the $k$-th client holds $n_k$ training data of  $\{(x_ {k, 1},y_{k,1})$, $(x_ {k, 2},y_ {k, 2})$, $\ldots, (x_ {k, n_k},y_ {k, n_k})\}$, the local objective function $L_ k(\cdot)$ can be defined as: 
\begin{equation}
	L_k(\mathbf{h}) = \frac{1}{n_k} \sum_{j=1}^{n_k} \mathcal{L}\left(\mathbf{h}(x_{k, j}) ,y_{k, j}\right),
\end{equation}
where $\mathcal{L}(,)$ denotes the loss function. Passing the aggregated global parameters to the client on the server side, and the $k$-th local client $\mathbf{h}_{k,e}^{t}$ performs $E$ steps local update: 
\begin{equation}
	\mathbf {h}_{k,e}^{t} \leftarrow \mathbf{h}_{k,e-1}^{t}-\eta^{t} \nabla L_ {k}\left(\mathbf{h}_{k,e-1}^{t}\right), e=1,\dots ,E,
\end{equation}
where $\eta^{t}$ denotes the learning rate, $t$ denotes the number of communications. Finally, global model aggregates the results of local training $\mathbf {h}_{1,E}^{t}, \ldots, \mathbf{h}_{K,E}^{t}$, and generates a new global model $\mathbf {h}^{t+1}$:
\begin{equation}
	\mathbf {h}^{t+1} = \sum_{k=1}^{K} \frac{n_k}{n} \mathbf{h}_{k,E}^{t}.
\end{equation}
where $n = \sum n_k$ denotes the total samples numbers of all clients.

\subsection{Generalization Bound Estimation} 

In this section, we incorporate distributional robustness analysis into the bound estimation of local model. In heterogeneous data scenarios, the use of upper and lower bounds provides a more robust and equitable measure of client training performance. This step is pivotal in our robust weighting aggregation strategy. Building upon \textbf{Theorem 3.1} for the upper bound and \textbf{Theorem 3.2} for the lower bound, we establish the following corollary:

\begin{corollary} The upper and lower bound of model performance, under the shifted distribution, are as follows: 
	\begin{equation}
		\begin{aligned}
			& Upper \; Bound  = \mathbf{E}_\mathbf{P}[\mathcal{L}^2(\mathbf{h}(X), Y)] \\
			& + 2 \lambda_\epsilon \left[\mathbf{V}_\mathbf{P}[\mathcal{L}^2(\mathbf{h}(X), Y)]\right]^{1/2} 
			+ \Delta^{u}_{\epsilon}, \label{prop1}
		\end{aligned}		
	\end{equation}
	\begin{equation*}
		\begin{aligned}
			& \Delta^{u}_{\epsilon} = 
			\epsilon^2\left(2-\epsilon^2\right)\left[M-\mathbf{E}_\mathbf{P}[\mathcal{L}^2(\mathbf{h}(X), Y)] \right. \\ 
			&  \left. -\frac{\mathbf{V}_\mathbf{P}[\mathcal{L}^2(\mathbf{h}(X), Y)]}{M^2-\mathbf{E}_\mathbf{P}[\mathcal{L}^2(\mathbf{h}(X), Y)]}\right].
		\end{aligned}
	\end{equation*}    
	\begin{equation}
		\begin{aligned}
			& Lower\; Bound = \mathbf{E}_\mathbf{P}[\mathcal{L}^2(\mathbf{h}(X), Y)] \\ 
			& -2 \lambda_\epsilon \left[\mathbf{V}_\mathbf{P}[\mathcal{L}^2(\mathbf{h}(X), Y)]\right]^{1/2} 
			- \Delta^{l}_{\epsilon},\label{prop2}
		\end{aligned}	
	\end{equation}
	$\Delta^{l}_{\epsilon} = \epsilon^2\left(2-\epsilon^2\right)\left[\mathbf{E}_\mathbf{P}[\mathcal{L}^2(\mathbf{h}(X), Y)]-\frac{\mathbf{V}_\mathbf{P}[\mathcal{L}^2(\mathbf{h}(X), Y)]}{\mathbf{E}_\mathbf{P}[\mathcal{L}^2(\mathbf{h}(X), Y)]}\right]$. 
\end{corollary}


\textbf{Corollary 4.1} outlines the upper and lower bounds for estimating the generalization performance based on the actual data distribution of each local client. These bounds primarily depend on the expectations and variances within the actual data distribution. Estimating these bounds involves sampling from the actual data distribution, considering training losses, and setting a predetermined distance to quantify the disagreement in data distribution.

\subsection{Robust Weighting for FL}

\begin{figure*}
	\centering
	\includegraphics[scale=0.05]{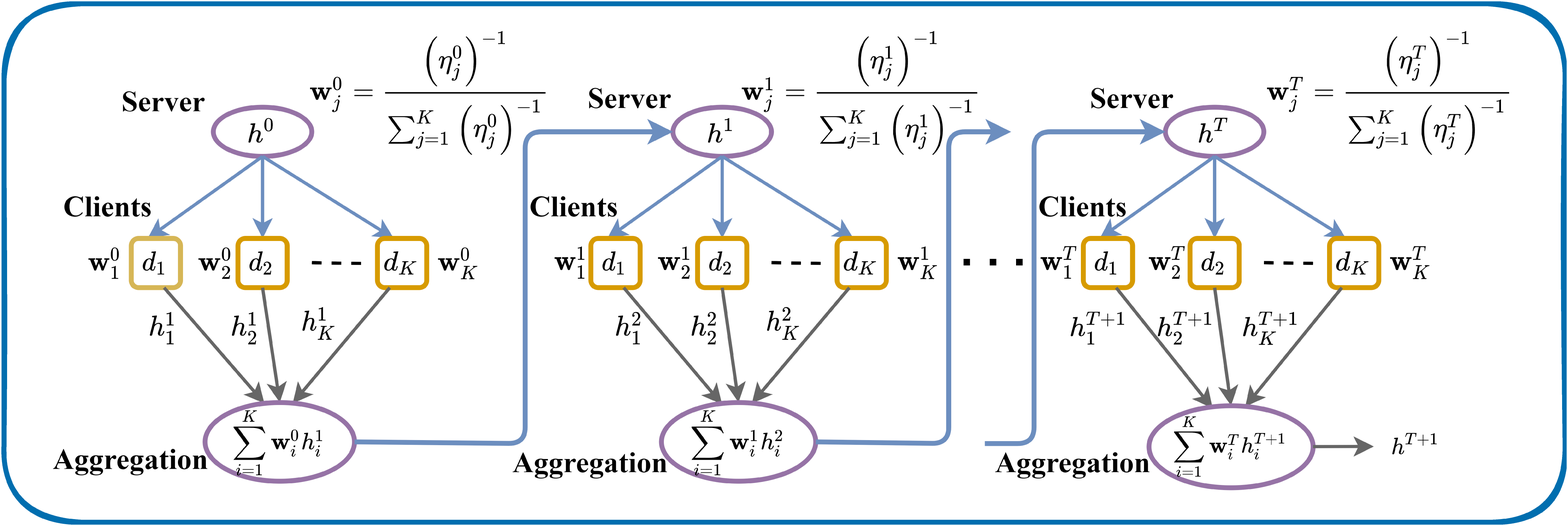}
	\caption{Overview of robust aggregation weighting. Each client estimates the model generalization disagreement, and then performs training and  aggregation weighting, $d_i$ represents the data distribution of the $i$-th ($i=1,\dots,K$) client.}
	\label{fig:flow chart}
\end{figure*}
In this section, we introduce our robust aggregation weighting strategy based on \textbf{Corollary 4.1}. Instead of relying solely on the samples proportion, our approach utilizes the disagreement in generalization bounds to achieve a more robust and equitable weighting scheme. In the context of Federated Learning, we assume that the data within each client remains unchanged during each round of training. However, different clients with diverse data distributions may exhibit varying generalization performances when subjected to the same model assumptions.

By denoting the upper bound as $\mathbf{u}$ and the lower bound as $\mathbf{l}$, as defined in equations (\ref{prop1}) and (\ref{prop2}) respectively, we calculate the bound disagreement $\mathbf{\sigma}$ for the $j$-th client at a given distance as follows:
\begin{equation}
	\mathbf{\sigma}_j^{(i)}= \Vert \mathbf{u}_j^{(i)} - \mathbf{l}_j^{(i)} \Vert , \label{sigma}
\end{equation}
Here, $\mathbf{\sigma}_j^{(i)}$ represents a point estimation at a specific distance $\delta$ within the range of $[0,\epsilon]$, where $\epsilon$ controls the maximum potential shift in the data distribution. In order to obtain more information regarding the disagreement in generalization bounds, we set multiple different distance values and compute their bound disagreements which can be viewed as neighborhood values. Therefor, the total bound disagreement $\eta_j$ of the $j$-th $(j=1,\dots,K)$ client is: 
\begin{equation} 
	\mathbf{\eta}_{j} = \sum_{i=1}^{\epsilon / \mathbf{\Delta} \delta} \mathbf{\sigma}_j^{(i)} , \label{eta}
\end{equation}
where $\mathbf{\Delta}\delta$ denotes a fixed interval. At the $t+1$-th round of aggregation weighting, the formulation of aggregated weighting as follows:
\begin{equation}
	\mathbf {h}^{t+1} \leftarrow \sum_{j=1}^K \frac{\left(1 / \eta_j^{t}\right)}{\sum_{j=1}^K\left(1 / \eta_j^{t}\right)} \mathbf {h}_j^{t+1}. \label{eta1}
\end{equation}
$\mathbf{w}_j^t = \frac{\left(1/\eta_j^{t}\right)}{\sum_{j=1}^K\left(1/\eta_j^{t}\right)}$ denotes the robust weighting. The framework of our robust weighting strategy is depicted in Figure \ref{fig:flow chart} and Algorithm \ref{alg}. It is important to note that the calculation of the upper and lower bounds in \textbf{Corollary 4.1} involves different conditions. As a result, direct formula inference and subtraction cannot be carried out, necessitating separate estimation of the upper and lower bounds.
\subsection{Robust Aggregation Weighting Algorithm}
\begin{algorithm} 
	\caption{: FL with Robust Aggregation Weighting.} 
	\label{alg} 
	\begin{algorithmic}[H]
		\STATE \textbf{Input:} Number of clients $K$, number of communications $T$.
		\STATE \textbf{Output:} The final global model $\mathbf {h}^{T+1}$.
		\STATE \textbf{ServerExecute:}
		\STATE initialize $\mathbf{h}^0$.
		\FOR { $t = 0,1,\cdots,T$}
		\FOR{ $j \in \{1,\cdots,K\}$ \;\textbf{in parallel} } 
		\STATE $\mathbf{h}_{j}^{t+1} \leftarrow$ \textbf{ClientUpdate}($j,\mathbf{h}^{t}$)  
		\STATE $\; \triangleright$ e.g.,FedAvg,FedProx,SCAFFOLD,FedDyn.  
		\STATE $\mathbf{\eta}_j^{t} = \sum \mathbf{\sigma}_i$, $i=1,\dots,\epsilon / \mathbf{\Delta} \delta.$ $\;\triangleright$ \textbf{Eq.~(\ref{sigma})}, bound disagreement estimation.
		\ENDFOR
		\STATE \textbf{Weightings Aggregation:}\\ $\mathbf {h}^{t+1} \leftarrow \sum_{j=1}^K \frac{\left( \eta_j^{t}\right)^{-1}}{\sum_{j=1}^K\left( \eta_j^{t}\right)^{-1}} \mathbf {h}_j^{t+1}.$  $\quad \triangleright$ \textbf{Eq.~(\ref{eta},\ref{eta1})}, robust aggregation weighting .		
		\ENDFOR
		\STATE return $\mathbf {h}^{T+1}$.		
	\end{algorithmic} 
\end{algorithm}
In this section, we abstract the key steps of the aforementioned algorithm and provide an overview of our algorithmic process. \textbf{Algorithm 1} presents the robust aggregation weighting strategy within a standard federated learning framework. The framework consists of two steps: \textbf{ClientUpdate} and \textbf{ServerExecute}. We also introduce four classical baselines that remain applicable in our algorithm settings. In the \textbf{ClientUpdate} step, the key difference lies in the estimation bound disagreements as weightings to aggregate the local model parameters, rather than using sample proportions. In the \textbf{ServerExecute} step, the server receives the disagreements and client models, and aggregates all the clients using new weightings that incorporate the estimated bound disagreements information.

\section{Experiments}

\textbf{Overview.} In this section, we conduct experiments to evaluate the capabilities of our robust aggregation weighting strategy. In Section \ref{E1}, we begin by verifying the significant differences in bound disagreement estimation between IID and Non-IID datasets. In Section \ref{E2}, we apply the bound disagreement estimation to weight the aggregation process in Federated Learning, utilizing four representative baseline algorithms as the backbone, namely FedAvg~\cite{mcmahan2017communication}, FedProx~\cite{li2020federated}, SCAFFOLD~\cite{pmlr-v119-karimireddy20a}, and FedDyn~\cite{acar2021federated}. These algorithms are evaluated on four datasets: MNIST, CIFAR10, CIFAR100 and EMNIST. Finally, in Section \ref{E3}, to observe the robustness of our strategy with different proportions of noisy data and clients, We set the noise ratio to 40\% on the CIFAR10 and EMNIST datasets. Additionally, under the same noise ratio as the baseline, we set the probability of client participation in training to 70\% on the CIFAR10 dataset.

\subsection{Bound Disagreements on IID and Non-IID Datasets}
\label{E1}

\begin{figure*}[!t]
	\centering
	\includegraphics[scale=0.5]{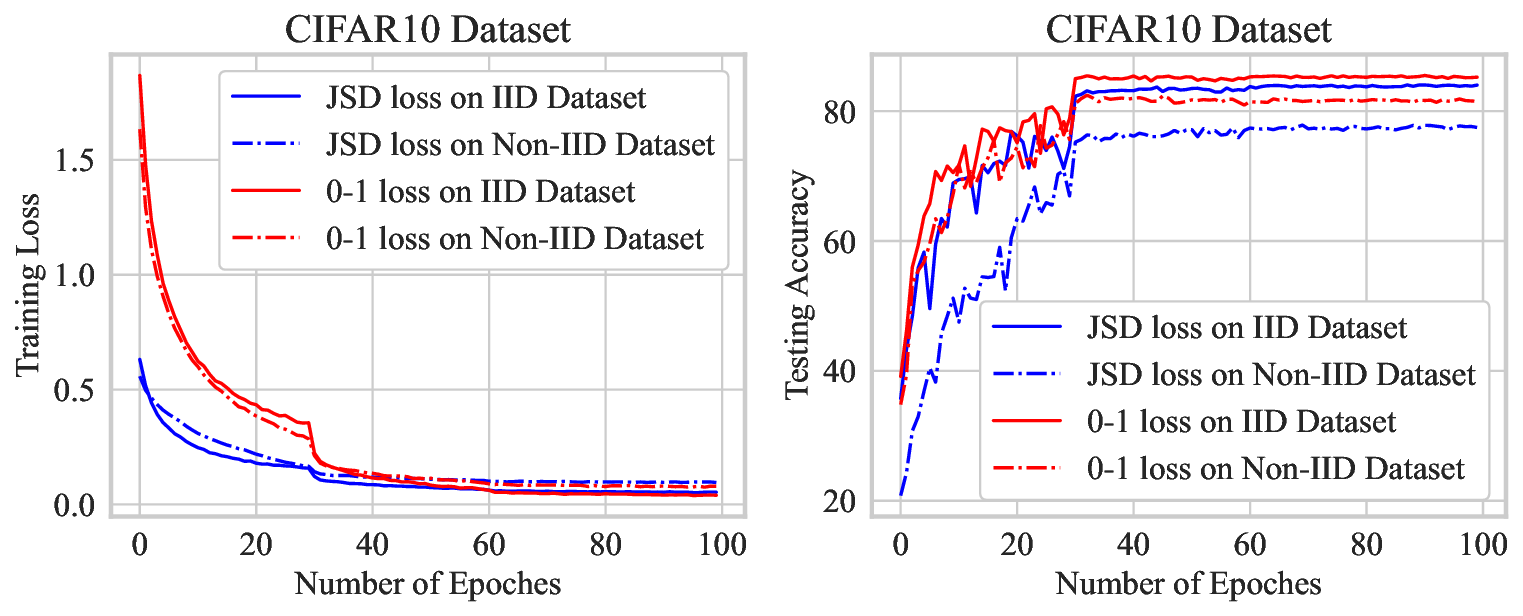}
	\caption{Testing accuracy and training loss on IID and Non-IID CIFAR10 dataset.}
	\label{fig:accuracy and loss}
\end{figure*}

\begin{figure*}[!t]
	\centering
	\includegraphics[scale=0.45]{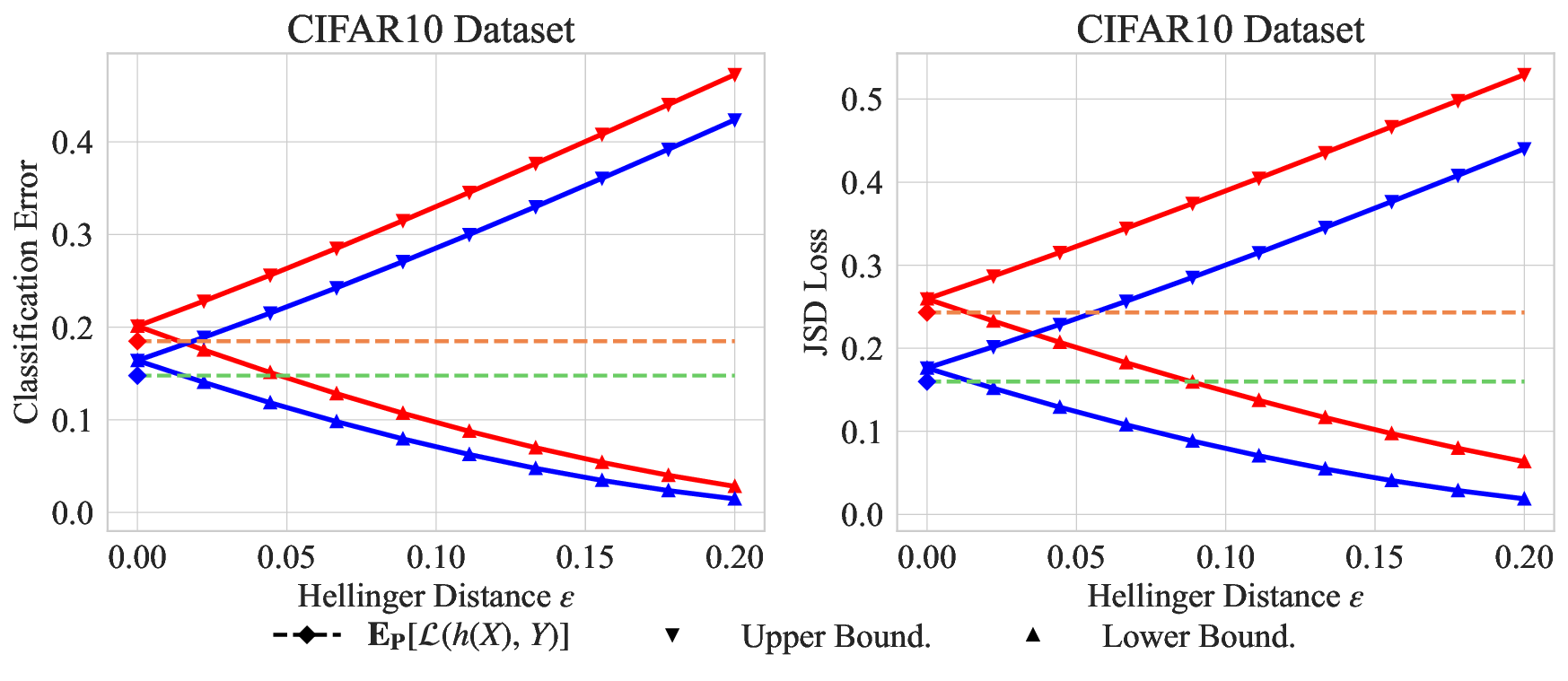}
	\caption{Upper and lower bounds on CIFAR10 dataset over 0-1 loss and JSD loss, the blue line represents IID data, the red line represents Non-IID data. We find that the bound disagreement of IID is tighter than Non-IID.}
	\label{fig:UL}
\end{figure*}

\subsubsection{Implementation}

This section investigates the differences in bound disagreement estimation on the CIFAR10 dataset for both the IID and Non-IID cases. For the IID case, the training dataset consists of 2,000 randomly selected samples from each category, resulting in a total of 20,000 training samples. For the Non-IID case, the training dataset also contains 20,000 samples, with each category having the following random sample sizes: [913, 994, 2254, 2007, 1829, 1144, 840, 4468, 713, 4838]. In all cases, the test set consists of 10,000 samples, with each category having 1,000 samples. We utilize the resnet20 network as the model and consider two loss functions: the 0-1 loss and JSD loss~\cite{NEURIPS2021_fe2d0103}. Throughout the experiments, we conduct 100 communication rounds with a batch size of 64.

\subsubsection{Results}

In Fig.~\ref{fig:accuracy and loss}, we illustrate the loss and test accuracy trends during training. The results clearly indicate that the model performs better on the IID case compared to the Non-IID case. Fig.~\ref{fig:UL} showcases the upper and lower bounds using the 0-1 loss and JSD loss and the blue line represents the IID data, while the red line represents the Non-IID data. We calculate the sum of bound disagreements at ten discrete points on a uniform interval. For the 0-1 loss, the total bound disagreement is 2.28 for the Non-IID data and 2.10 for the IID data. In the case of the JSD loss, the total bound disagreement is 2.38 for the Non-IID data and 2.16 for the IID data. These findings indicate that in the IID scenario, where the data are more similar to each other and follow the same distribution, the range of possible prediction outcomes is smaller, resulting in a tighter bound on the disagreement between different models or algorithms. Conversely, in the Non-IID scenario, where the data are more diverse and may follow different distributions, the range of possible prediction outcomes is wider, leading to a looser bound on the disagreement between models.

These results inspire us to recognize that the shift in heterogeneous data distribution can be effectively assessed through the estimation of its generalization performance bounds. Furthermore, they provide a preliminary understanding for subsequent experiments on robust aggregation weighting.
\subsection{Robust Aggregation Weighting for FL }
\label{E2}
\begin{table*}
	\caption{Test accuracy with $Dirichlet(0.3,0.9)$,including 10, 20, 50, 100, and 200 clients.}
	\label{tabel1}
	\centering
	\begin{tabular}{c|c|c|cccc|cccc}
		\toprule
		Backbone & Clients & Act.prob  & FedAvg & FedProx &SCAFFOLD & FedDyn & FedAvg & FedProx &SCAFFOLD & FedDyn\\		
		\midrule
		Weighting Strategy   &\multicolumn{2}{c|}{Configuration}  &\multicolumn{4}{c|}{Test Accuracy on MNIST(\%)} &\multicolumn{4}{c}{Test Accuracy on CIFAR10 (\%)}\\ 
		\cmidrule(lr){1-11}
		Propto. & 10 & $1.0$ &  72.21& 71.17 & 87.74 & 90.97  & 64.50 &65.14 &77.62 &79.51  \\
		Robust. & 10 & $1.0$ & \textbf{96.86} $\uparrow$ & \textbf{96.84} $\uparrow$ & \textbf{95.50} $\uparrow$ & \textbf{96.60} $\uparrow$ & \textbf{77.37} $\uparrow$ & \textbf{77.63} $\uparrow$ & \textbf{78.18} $\uparrow$ & \textbf{79.99} $\uparrow$  \\ 
		\cmidrule(lr){1-11}
		Propto. & 20 & $1.0$ & 80.13 & 81.00  & 77.89 & 93.76 & 64.60 & 64.35  & 62.33 &79.49  \\
		Robust. & 20 & $1.0$ & \textbf{95.63} $\uparrow$ & \textbf{95.33} $\uparrow$ & \textbf{89.18} $\uparrow$ & \textbf{95.25} $\uparrow$ & \textbf{77.84}$ \uparrow$ & \textbf{78.38} $\uparrow$ & \textbf{68.94} $\uparrow$ & \textbf{80.78} $\uparrow$  \\
		\cmidrule(lr){1-11}
		Propto. & 50 & $1.0$ & 86.95  &86.63  &96.89  &95.58 & 61.19  &60.97  &59.40  &78.47 \\
		Robust. & 50 & $1.0$ & \textbf{97.33} $\uparrow$ & \textbf{97.31} $\uparrow$ & \textbf{97.29} $\uparrow$ & \textbf{96.78} $\uparrow$ & \textbf{73.48} $\uparrow$ &  \textbf{73.32} $\uparrow$& \textbf{73.64} $\uparrow$ & \textbf{78.86} $\uparrow$ \\
		\cmidrule(lr){1-11}
		Propto. & 100 & $1.0$ & 90.15  & 89.91 & 97.27 &95.85 & 62.14  &62.59  &74.23  & 76.51   \\   
		Robust. & 100 & $1.0$ & \textbf{97.32} $\uparrow$  & \textbf{97.28} $\uparrow$ & \textbf{97.51} $\uparrow$ & \textbf{96.73} $\uparrow$ & \textbf{70.98} $\uparrow$ & \textbf{71.13} $\uparrow$ & \textbf{72.70}  & \textbf{77.02} $\uparrow$ \\
		\cmidrule(lr){1-11}
		Propto. & 200 & $1.0$ & 90.66  &90.72  &97.25  &95.58 & 63.06  &63.35  &72.82  &76.15  \\  
		Robust. & 200 & $1.0$  & \textbf{96.70} $\uparrow$ & \textbf{96.68} $\uparrow$ & \textbf{97.33} $\uparrow$  & \textbf{96.25} $\uparrow$  & \textbf{70.71} $\uparrow$ & \textbf{71.05} $\uparrow$ &\textbf{72.10}  & \textbf{76.41} $\uparrow$\\    		    				    		
		\midrule
		Weighting Strategy   &\multicolumn{2}{c|}{Configuration} &\multicolumn{4}{c|}{Test Accuracy on CIFAR100 (\%)} &\multicolumn{4}{c}{Test Accuracy on EMNIST (\%)}\\  	   
		\cmidrule(lr){1-11}
		Propto. & 10  & $1.0$ & 37.37  &36.97 &37.94 &47.48  &63.75  &64.17 &85.84 &89.32\\
		Robust. & 10  & $1.0$ &\textbf{45.76} $\uparrow$ & \textbf{45.65} $\uparrow$ & \textbf{40.07} $\uparrow$ & \textbf{47.00} &\textbf{88.54} $\uparrow$ & \textbf{89.43} $\uparrow$ & \textbf{87.06} $\uparrow$ & \textbf{90.52} $\uparrow$  \\ 
		\cmidrule(lr){1-11}
		Propto. & 20  &$1.0$ & 32.48  &32.76  &44.26 &49.12  &71.09  &69.51  &58.96 &90.52 \\
		Robust. & 20  &$1.0$ &\textbf{44.51}$ \uparrow$ & \textbf{44.52} $\uparrow$ & \textbf{43.42}  & \textbf{48.80}  & \textbf{90.73}$ \uparrow$ & \textbf{91.13} $\uparrow$ & \textbf{61.90} $\uparrow$ & \textbf{91.23} $\uparrow$ \\
		\cmidrule(lr){1-11}
		Propto. & 50  &$1.0$ & 31.88  &32.02  &41.23  &49.42  &80.05  &80.81  &91.67  &90.47 \\
		Robust. & 50  &$1.0$ & \textbf{41.32} $\uparrow$ &  \textbf{41.19} $\uparrow$& \textbf{41.77} $\uparrow$& \textbf{47.75} & \textbf{91.94} $\uparrow$ &  \textbf{91.92} $\uparrow$& \textbf{92.05} $\uparrow$& \textbf{91.19} $\uparrow$ \\
		\cmidrule(lr){1-11}
		Propto. & 100  &$1.0$ & 28.65  &29.35  &40.53  &47.46 &81.32  &80.70  &75.67  &90.60 \\  
		Robust. & 100  &$1.0$ & \textbf{39.48} $\uparrow$ & \textbf{39.20} $\uparrow$ &\textbf{42.52} $\uparrow$ & \textbf{47.04} &\textbf{92.03} $\uparrow$ & \textbf{92.10} $\uparrow$ & \textbf{91.91} $\uparrow$& \textbf{90.27}  \\
		\cmidrule(lr){1-11}  
		Propto. & 200  & $1.0$ & 28.79  &28.40  &35.18  &44.78 &83.13  &83.26  &90.91  & 90.49  \\  
		Robust. & 200  & $1.0$ & \textbf{34.77} $\uparrow$ & \textbf{34.36} $\uparrow$ & \textbf{37.40} $\uparrow$  & \textbf{44.01}  &\textbf{91.79} $\uparrow$ & \textbf{91.80} $\uparrow$ &  \textbf{92.55} $\uparrow$ & \textbf{90.83} $\uparrow$ \\  		    		   	 		    	
		\bottomrule
	\end{tabular}
\end{table*}

In this section, our experiments are to verify the effectiveness of robust aggregation weighting by using bound disagreements in FL. The baselines that are selected follow the paper of FedDyn~\cite{acar2021federated}, specifically including FedAvg~\cite{mcmahan2017communication}, FedProx~\cite{li2020federated}, SCAFFOLD~\cite{pmlr-v119-karimireddy20a}, and FedDyn~\cite{acar2021federated}. Under the same hyperparameter setting, we compare the proportion of the sample weighting with our robust aggregation weighting. 

\subsubsection{Experimental Settings}

\textbf{Datasets.} To evaluate data heterogeneity, we utilize four widely-used datasets in Federated Learning research: CIFAR10, MNIST, CIFAR100, and EMNIST. In order to create a more realistic simulation of a Non-IID dataset, we introduce non-uniform distributions to the clients' classes and allow for the possibility of some classes being missing. For this purpose, we sample from a non-equilibrium Dirichlet distribution~\cite{acar2021federated}. For each client, we generate a random vector  $\mathbf{p}_k \sim \textbf{Dir}(\alpha)$  from the Dirichlet distribution, where $\mathbf{p}_k=\left[p_{k, 1}, \ldots, p_{k,C}\right]^{\top}$. The proportion of images belonging to each category $c$ in the dataset allocated to the $k$-th client is represented by $(100 \cdot p_{k,c})\%$. In our experiments, we set the parameter of the lognormal distribution to unbalanced\_sgm = 0.9 and the parameter of the Dirichlet distribution to rule\_arg = 0.3. Additionally, to simulate noisy data in a real-world scenario, we introduce 20\% noisy data to the four datasets by assigning some of the labels as 0.

\begin{figure*}
	\centering
	\subfigure[Test accuracy with 10 clients on four datasets.]{		
		\includegraphics[width=0.9\textwidth]{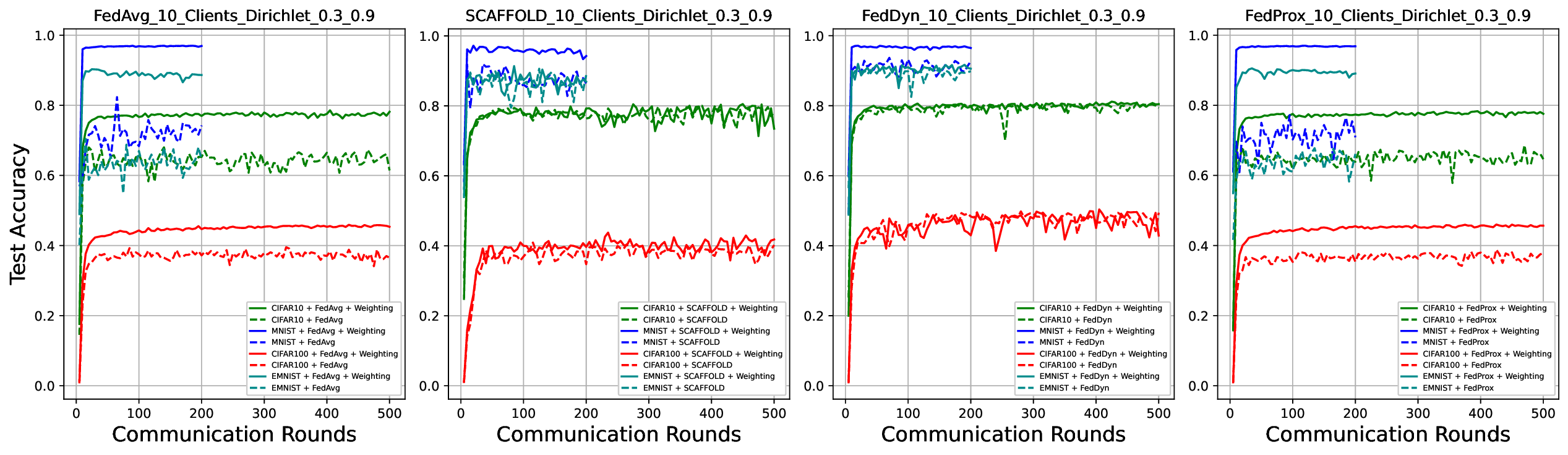}
	}
	\subfigure[Test accuracy with 20 clients on four datasets.]{
		\includegraphics[width=0.9\textwidth]{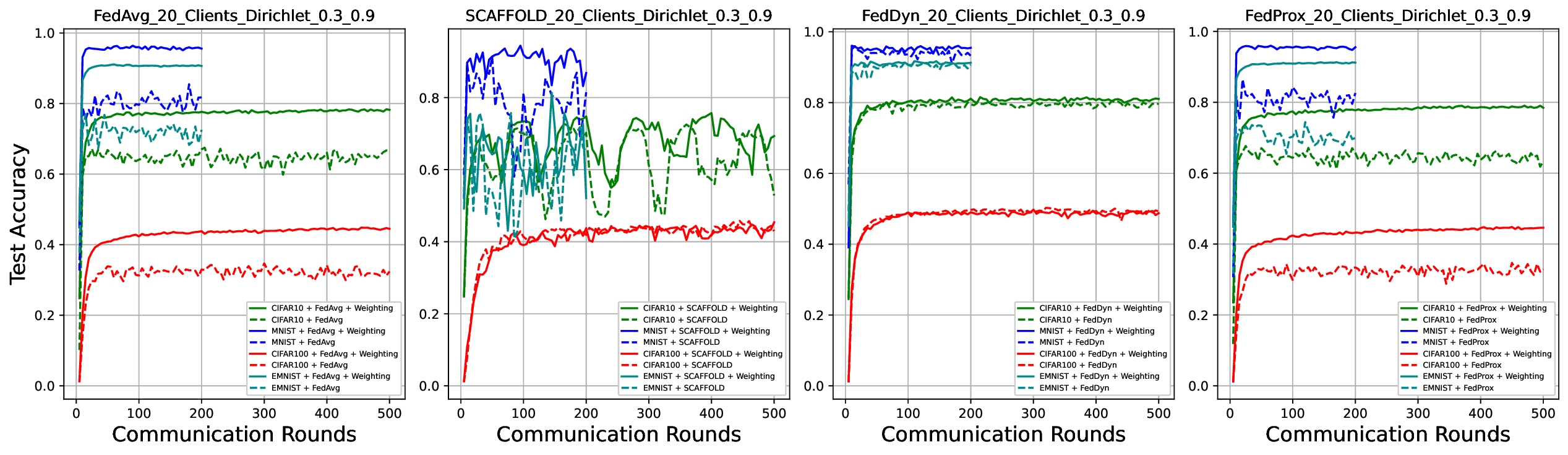}
	}
	\hspace{0.3in}
	\subfigure[Test accuracy with 50 clients on four datasets.]{
		\includegraphics[width=0.9\textwidth]{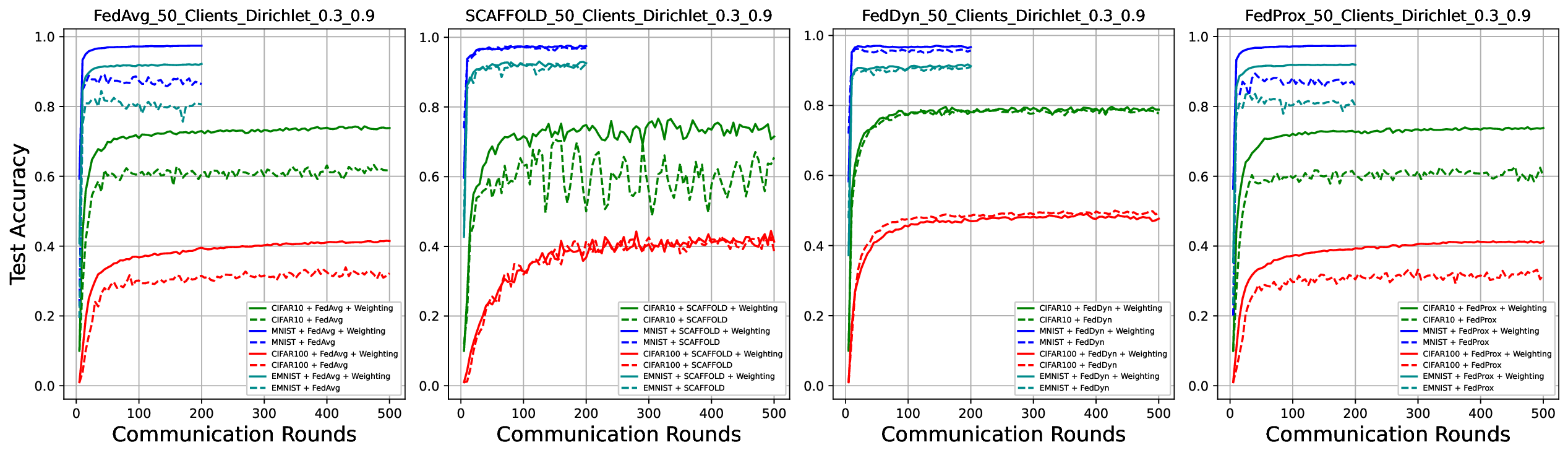}
	}
	\caption{CIFAR10, MNIST, CIFAR100, EMNIST with $Dirichlet(0.3,0.9)$.  Communication rounds are set to [200,500,700] for different datasets.}
	\label{fig:datasets_dirichlet}
\end{figure*}

\textbf{Settings.} For all experiments, we assume that all clients participate in each round of communication, i.e., the probability of each client participating in the training is equal to 1. The number of communication rounds is set to [200,500,700] for different datasets as displayed in Fig.~\ref{fig:datasets_dirichlet} . The weight decay is equal to 1e-3, and the batch size is 50.  In each client, the local epoch is  5, and the learning rate is 0.1.  On each dataset, we conduct experiments on 10, 20, 50, 100 and 200 clients, respectively. For the MNIST and EMNIST datasets, we use a fully connected neural network consisting of two hidden layers with the number of neurons in the hidden layers being 200 and 100 respectively. For the experiment of CIFAR10 and CIFAR100 datasets, we use a CNN model used in (McMahan et al., 2017)~\cite{mcmahan2017communication},  including 2 convolution layers and 64 $\times$ 5 $\times$ 5 filters, followed by 2 fully connected layers with 394 and 192 neurons and a softmax layer.

\subsubsection{Experimental Results on Model Performance}

 \textbf{Overall.} We apply both the robust aggregation weighting strategy and the original samples proportion method to the four classical baselines. Fig.~\ref{fig:datasets_dirichlet} illustrates the test accuracy results obtained using 10, 20 and 50 clients, with the addition of 20\% noisy data. In the figure, the solid lines represent the robust aggregation weighting strategy, while the dashed lines represent the samples proportion strategy. The corresponding test accuracies for all experiments are provided in Table~\ref{tabel1}. In Table~\ref{tabel1}, the term \textbf{Propto} indicates the proportion of the samples, and \textbf{Robust} represents the robust aggregation weighting. From the table, it is evident that our strategy consistently achieves higher test accuracies compared to the original strategy. Our experiments demonstrate significant performance improvements over FedAvg and FedProx, as well as a slight improvement over SCAFFOLD and FedDyn. These results indicate that our robust aggregation weightings are more fair and robust when dealing with heterogeneous and noisy data.

\textbf{Test Accuracy.} Specifically, by observing the experimental data on federated learning test accuracy at Table~\ref{tabel1}, we can see that our weighting method consistently achieves significant improvements compared to the baseline method in FedAvg and FedProx. Additionally, it can be observed that under the presence of a higher proportion of noisy data and strong heterogeneity in the data distribution among clients, FedProx, as a regularization term improvement over FedAvg, does not outperform the FedAvg method in a stable manner. In SCAFFOLD and FedDyn, when considering all the clients and datasets, overall 77.5\% of the test results outperform the original method. On the MNIST dataset, the test accuracy is superior to the original method entirely, followed by a failure rate of 2.5\% on EMNIST, 5\% on CIFAR10, and finally 15\% on CIFAR100. We attribute this phenomenon to the dimension collapse~\cite{shi2023towards} caused by the heterogeneity of dataset in the training process when introducing bias correction for weighting estimation, leading to the loss of some representation information and resulting in false ineffective weightings, thereby reducing the model's performance. For the FedDyn method with outstanding performance, even with the introduction of new weightings in SCAFFOLD, FedProx, and FedAvg, it remains challenging to match the performance of the original weighted model. FedDyn's strategy to converge the model on the client side towards the global optimum makes it challenging to capture effective bound disagreements on certain well-performing clients, but it is still an excellent solution. 

\textbf{Robustness Analysis.} In addition to the aforementioned analysis, we conducted a robustness analysis to evaluate the performance of our robust aggregation weighting strategy. The experimental results demonstrate that our robust aggregation weighting strategy promotes fairness by assigning smaller weightings to clients with a high percentage of heterogeneous data during parameter aggregation. Notably, the robust aggregation weighting strategy yields significant accuracy improvements in the case of FedAvg and FedProx, while offering a slight improvement for SCAFFOLD and FedDyn. The reason behind these results lies in the fact that FedAvg and FedProx are classical methods that do not adequately address the correction of local model bias caused by heterogeneous data. In contrast, SCAFFOLD and FedDyn focus on optimizing the local model shift resulting from heterogeneous data. Although SCAFFOLD and FedDyn achieve improved accuracy, they require a higher number of training time compared to FedAvg and FedProx. By incorporating our robust aggregation weighting strategy, FedAvg and FedProx achieve comparable performance to SCAFFOLD and FedDyn while requiring fewer computation time.

\textbf{Communication and Performance Balance.} From a horizontal perspective, the differences in test accuracy between our new weighting method and the original method are narrowing across the four baselines. For example, when comparing FedAvg and FedProx with SCAFFOLD, in experiments where the weightings are based on sample proportions, FedAvg and FedProx consistently perform significantly worse than SCAFFOLD. However, in experiments where the weightings are based on bound disagreements, the performance differences between the algorithms rapidly diminish, and in a small number of cases, FedAvg and FedProx outperform SCAFFOLD. Similarly, although FedAvg and FedProx slightly trail behind FedDyn in terms of model performance after incorporating bound information for  aggregation weightings, the differences are significantly smaller compared to the sample proportion weighting method. This is particularly applicable in scenarios where there are strict communication and computation constraints but more relaxed accuracy requirements.

Furthermore, to assess the stability of the experiments, we conducted additional experiments involving 100 and 200 clients, as shown in Fig.~\ref{fig:datasets_dirichlet_addtional}. The results obtained in these experiments are consistent with the improvements observed in the 10, 20 and 50 client experiments. More detailed experimental results are also presented in Table~\ref{tabel1}. In Table~\ref{tabel1}, our experimental results demonstrate notable accuracy improvements for FedAvg and FedProx. However, it is worth noting that our weighting strategy does not consistently enhance the accuracy for all federated learning algorithms, such as SCAFFOLD and FedDyn. This variability in performance may be attributed to the stability of these FL algorithms and the potential risk of training overfitting. Overall, the experimental results provide compelling evidence for the effectiveness of our weighting strategy, particularly in the case of FedAvg and FedProx. 

\subsubsection{Analysis on Variance of Test Accuracy }

In this section, we conduct experiments on four baselines and four datasets to randomly select 10 and 100 client counts and calculate the variance of the test accuracy after model convergence, as shown in Table~\ref{table_v}. Similar to the previous table settings, \textbf{Propto.} represents the sample proportion weighting method, while \textbf{Robust.} represents our new method based on generalization bound estimation. The variance is computed based on the performance of the models that continue to participate in communication training after convergence. We conduct this analysis because we find that in environments with high noise ratios and strong data heterogeneity, some methods such as SCAFFOLD~\ref{fig:datasets_dirichlet} exhibit significant oscillations after convergence, which increases the uncertainty of model training.

\begin{figure*}
	\centering	
	\subfigure[Test accuracy with 100 clients on four Datasets]{
		\includegraphics[width=0.9\textwidth]{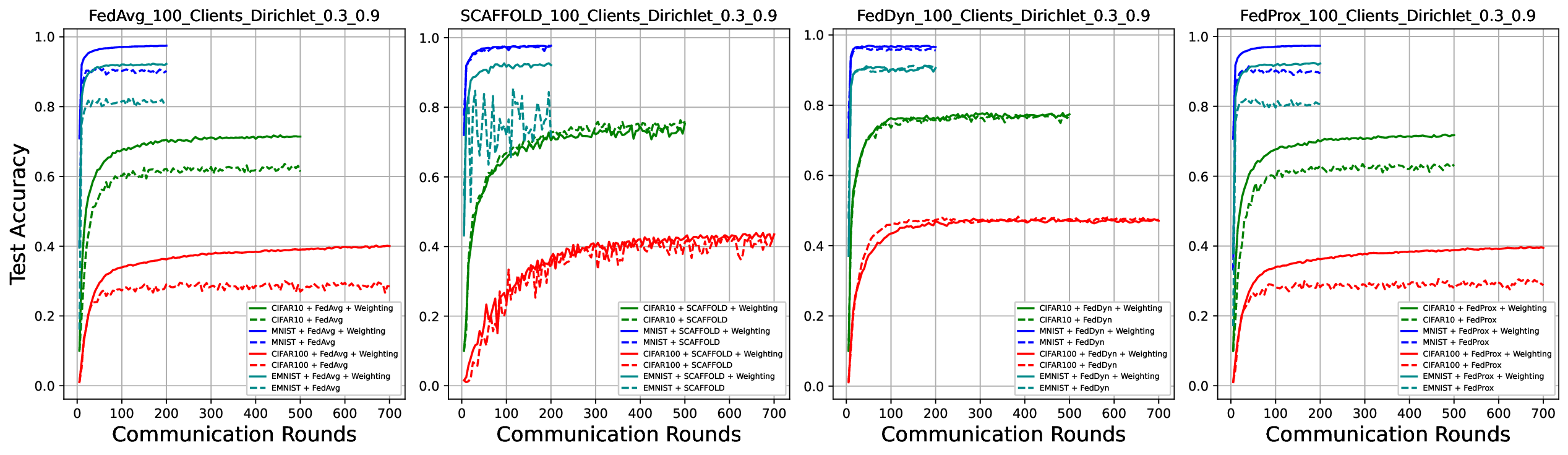}
	}
	\hspace{0.3in}
	\subfigure[Test accuracy with 200 clients on four Datasets]{
		\includegraphics[width=0.9\textwidth]{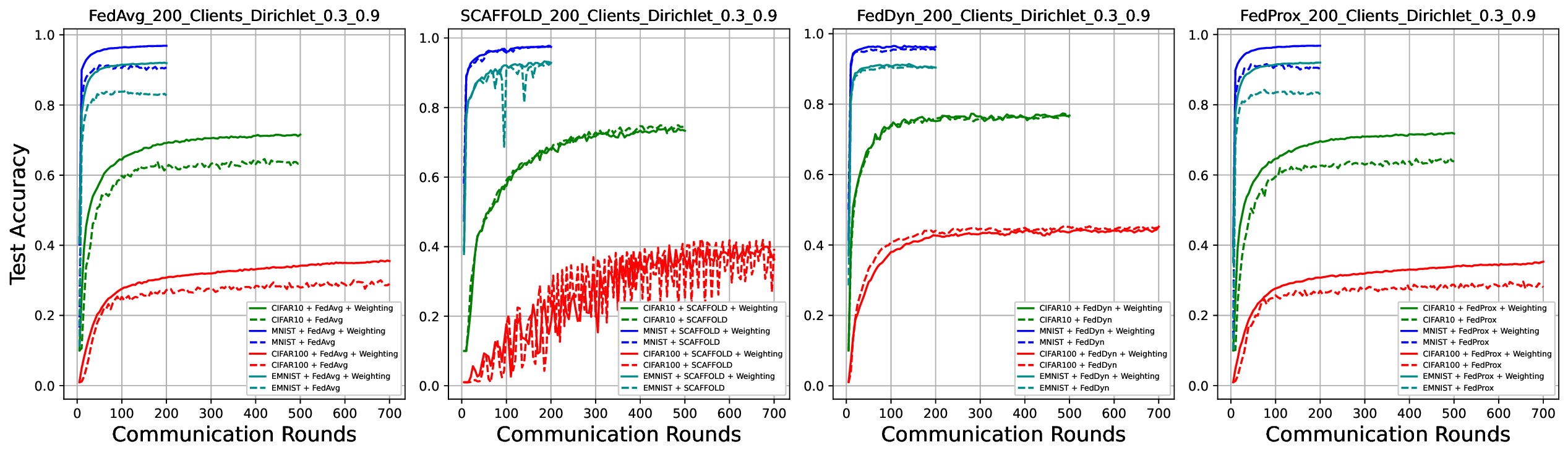}
	}
	\caption{CIFAR10, MNIST, CIFAR100, EMNIST with $Dirichlet(0.3,0.9)$.}
	\label{fig:datasets_dirichlet_addtional}
\end{figure*}

Therefore, we compare the performance stability of several baselines, which can be visually observed in Fig.~\ref{fig:datasets_dirichlet} and Fig.~\ref{fig:datasets_dirichlet_addtional}. Specifically, from Table~\ref{table_v}, we find that our weighting method overall had a lower test accuracy variance than the original weighting method in 90.625\% of cases. The test accuracy variances of FedAvg and FedProx methods were reduced when using our proposed weighting scheme, compared to the original sample proportion weightings. In the case of FedDyn and SCAFFOLD methods, only a small portion of the variances exceeded those of the original weighting method. This indicates that our method achieves robust training in the majority of cases, but it may fail in some extreme scenarios, which requires further exploration.
\begin{table}
	\setlength{\tabcolsep}{4pt}
	\caption{Variance about test accuracy with $Dirichlet(0.3,0.9)$ }
	\label{table_v}
	\centering
	\begin{tabular}{cccccc}
		\toprule
		Backbone  & Clients  & FedAvg & FedProx &SCAFFOLD & FedDyn\\
		\midrule
		&\multicolumn{5}{c}{Variance about Test Accuracy on CIFAR10} \\  
		\cmidrule(lr){2-6}
		Propto. & 10  &0.0180  &0.0186  &0.0148  &0.0113   \\
		Robust. & 10  & \textbf{0.0039} $\downarrow$  & \textbf{0.0037} $\downarrow$ & \textbf{0.0194}  & \textbf{0.0074} $\downarrow$  \\
		\cmidrule(lr){2-6}
		Propto. & 100   &0.0059  &0.0064  &0.0106  &0.0056 \\  
		Robust. & 100 & \textbf{0.0043} $\downarrow$  & \textbf{0.0044} $\downarrow$  & \textbf{0.0106}  & \textbf{0.0045} $\downarrow$  \\
        \midrule
		&\multicolumn{5}{c}{Test Accuracy on MNIST} \\
		\cmidrule(lr){2-6}
		Propto. & 10  &0.0307  &0.0307  &0.0239  &0.0181   \\
		Robust. & 10 & \textbf{0.0010} $\downarrow$ & \textbf{0.0009} $\downarrow$ & \textbf{0.0084} $\downarrow$  & \textbf{0.0029} $\downarrow$ \\
		\cmidrule(lr){2-6}
		Propto. & 100   &0.0041  &0.0038  &0.0041  &0.0020 \\  
		Robust. & 100 & \textbf{0.0009} $\downarrow$ & \textbf{0.0007} $\downarrow$  & \textbf{0.0010} $\downarrow$ & \textbf{0.0010} $\downarrow$  \\
		\midrule
		&\multicolumn{5}{c}{Test Accuracy on EMNIST} \\
		\cmidrule(lr){2-6}
		Propto. & 10  &0.0234  &0.0251  &0.0241  &0.0134   \\
		Robust. & 10 &\textbf{0.0065} $\downarrow$ & \textbf{0.0050} $\downarrow$  & \textbf{0.0180} $\downarrow$ & \textbf{0.0101} $\downarrow$  \\
		\cmidrule(lr){2-6}
		Propto. & 100   &0.0057  &0.0065  &0.0596  & 0.0042\\  
		Robust. & 100 & \textbf{0.0015} $\downarrow$ & \textbf{0.00321} $\downarrow$ & \textbf{0.0059}$\downarrow$ & \textbf{0.0043}   \\ 
        \midrule
		&\multicolumn{5}{c}{Test Accuracy on CIFAR100} \\
		\cmidrule(lr){2-6}
		Propto. & 10  &0.0098  &0.0095  &0.0173  &0.0110   \\
		Robust. & 10 & \textbf{0.0027} $\downarrow$ & \textbf{0.0023} $\downarrow$  & \textbf{0.0146} $\downarrow$ & \textbf{0.0248}  \\
		\cmidrule(lr){2-6}
		Propto. & 100   &0.0070  &0.0065  &0.0165  &0.0030 \\  
		Robust. & 100 & \textbf{0.0038} $\downarrow$ & \textbf{0.0031}$\downarrow$  & \textbf{0.0082}$\downarrow$ & \textbf{0.0026} $\downarrow$  \\
		\bottomrule
	\end{tabular}
\end{table} 

\subsection{Different Proportion of Noisy Data and Clients in FL}
\label{E3}
\begin{figure*}[!t]
	\centering
	\subfigure[Test accuracy with on 40\% noise data.]{		
		\includegraphics[width=0.9\textwidth]{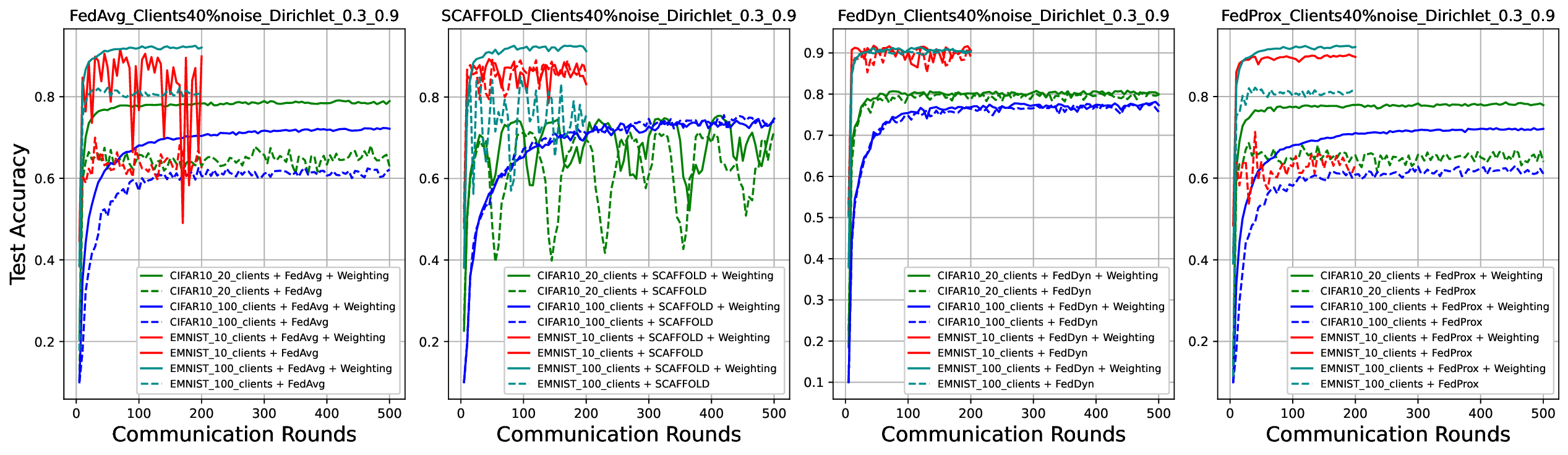}
	}
	\hspace{0.3in}
	\subfigure[Test accuracy with 70\% clients involved in training.]{
		\includegraphics[width=0.9\textwidth]{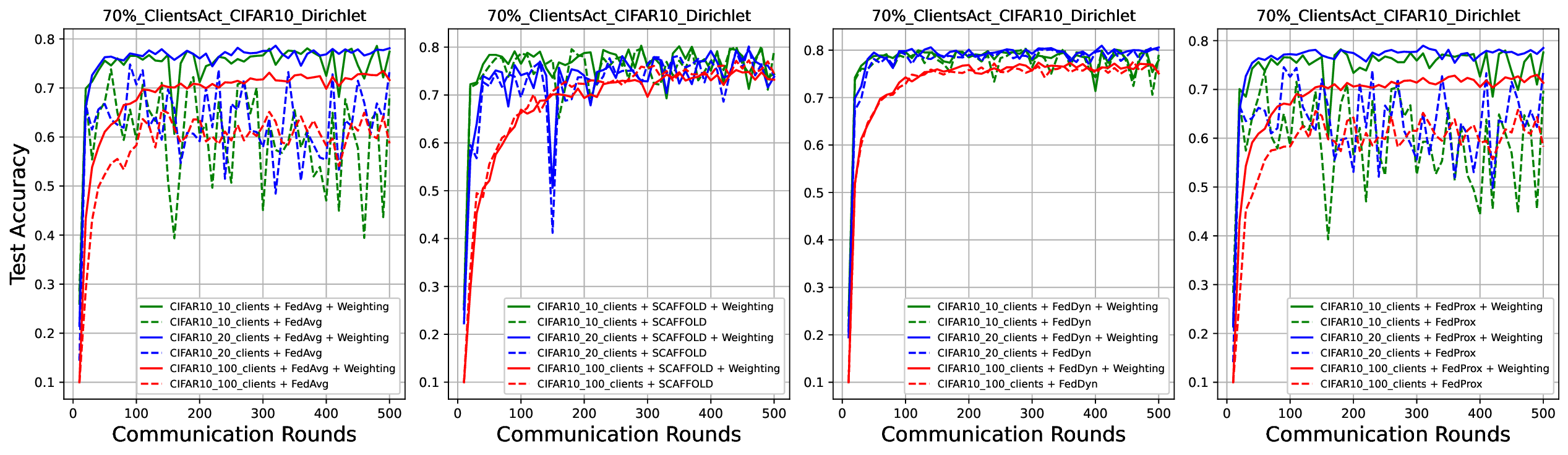}
	}
	\caption{(a) Test accuracy on CIFAR10 with 40\% data noise and $Dirichlet(0.3,0.9)$. (b) Test Accuracy on CIFAR10 with 70\% clients involved in training and $Dirichlet(0.3,0.9)$. The solid line represents our weighting strategy, while the dashed line represents the use of training sample proportion as weights.}
	\label{fig:10_40noiseCIFAR10_dirichlet}
\end{figure*}

\begin{table}
	\setlength{\tabcolsep}{4pt}
	\caption{Test accuracy with $Dirichlet(0.3,0.9)$ on 40\% noise data.}
	\label{table2}
	\centering
	\begin{tabular}{cccccc}
		\toprule
		Backbone  & Clients  & FedAvg & FedProx &SCAFFOLD & FedDyn\\
		\midrule
		&\multicolumn{5}{c}{Test Accuracy on CIFAR10(\%)} \\  
		\cmidrule(lr){2-6}
		Propto. & 20  & 64.87 & 65.06 & 62.91 & 79.56  \\
		Robust. & 20  & \textbf{78.48}$ \uparrow$ & \textbf{77.89} $\uparrow$ & \textbf{68.84} $\uparrow$ & \textbf{80.26} $\uparrow$  \\
		\cmidrule(lr){2-6}
		Propto. & 100   &61.35  &61.60  &74.24  &77.06 \\  
		Robust. & 100 & \textbf{72.13} $\uparrow$ & \textbf{72.00} $\uparrow$ & \textbf{73.38}  & \textbf{77.41} $\uparrow$  \\
		\midrule
		&\multicolumn{5}{c}{Test Accuracy on EMNIST(\%)} \\
		\cmidrule(lr){2-6}
		Propto. & 10  &63.35  &63.66  &86.40 &89.44  \\
		Robust. & 10 & \textbf{81.72}$ \uparrow$ & \textbf{89.74} $\uparrow$ & \textbf{85.66}  & \textbf{90.24} $\uparrow$  \\
		\cmidrule(lr){2-6}
		Propto. & 100   &80.81  &80.88  &75.23  &90.26  \\  
		Robust. & 100 & \textbf{92.09} $\uparrow$ & \textbf{92.11} $\uparrow$ & \textbf{92.03} $\uparrow$& \textbf{90.70} $\uparrow$  \\    		    	
		\bottomrule
	\end{tabular}
\end{table} 
In this section, we conducted experiments to verify the effectiveness of our robust aggregation weighting strategy under random selected proportions of noisy data and varying proportions of participating clients. To introduce noise, we added 40\% noise to both the CIFAR10 and EMNIST datasets. For proportions of participating clients, we set the proportion of clients participating in the training on the CIFAR-10 dataset to act\_prob=0.7. The rest of the experimental setup remains the same. The results of these experiments are summarized in Fig.~\ref{fig:10_40noiseCIFAR10_dirichlet}, Table~\ref{table2} and Table~\ref{table3}. By comparing the results from the aforementioned experiments, we observed that our robust aggregation weighting strategy remains effective even when dealing with a higher proportion of noisy data or when only a subset of clients participate in the training. The strategy consistently improves the test accuracy across different scenarios, demonstrating its robustness and adaptive nature.

These findings highlight the robustness of our aggregation weighting strategy in the presence of highly noisy data. It showcases its effectiveness even when a substantial portion of the data is corrupted and when only a subset of clients contribute to the training process. The ability of our strategy to adapt to such challenging scenarios is a significant advantage, as it ensures reliable and accurate model training in real-world settings.
\begin{table}
	\setlength{\tabcolsep}{4pt}
	\caption{Test accuracy with $Dirichlet(0.3,0.9)$ on 70\% clients.}
	\label{table3}
	\centering
	\begin{tabular}{ccc|ccc}
		\toprule
		Backbone & Strategy & Act.prob & 10 & 20  & 100 \\		
		\midrule
		& \multicolumn{2}{c}{Configuration} &\multicolumn{3}{c}{Test Accuracy on CIFAR10(\%)}\\
		\cmidrule(lr){2-6}		
		\multirow{2}{*}{FedAvg} & Propto. & $0.7$ &62.32  &63.59  &61.71   \\
		& Robust. & $0.7$& \textbf{75.92}$ \uparrow$ & \textbf{77.15} $\uparrow$ & \textbf{72.01} $\uparrow$ \\				
		\cmidrule(lr){2-6}		
		\multirow{2}{*}{FedProx} & Propto. & $0.7$  &77.23  & 73.48 &74.46   \\
		& Robust. & $0.7$ & \textbf{77.50} $\uparrow$ & \textbf{75.49} $\uparrow$  & \textbf{73.24} \\			
		\cmidrule(lr){2-6}
		\multirow{2}{*}{SCAFFOLD} & Propto. &$0.7$   & 78.10 & 79.00 & 76.13\\
		& Robust. & $0.7$  & \textbf{78.58}$ \uparrow$ & \textbf{79.70} $\uparrow$ & \textbf{76.58}$\uparrow$  \\				
		\cmidrule(lr){2-6}
		\multirow{2}{*}{FedDyn} & Propto. & $0.7$   &61.22  & 64.27 &61.78 \\
		& Robust. & $0.7$ & \textbf{75.82} $\uparrow$ & \textbf{77.54} $\uparrow$ & \textbf{71.77}$\uparrow$ \\
		\bottomrule
	\end{tabular}
\end{table}

\section{Conclusion}

In the context of Federated Learning (FL), it has come to our attention that the traditional approach of assigning aggregation weights based on sample proportions may result in unfairness due to the widespread statistical heterogeneity among local models. To address the potential negative effects stemming from this issue, we have reconsidered the aggregation weighting method from a novel perspective, specifically by taking into account the generalization performance of each local client model. Inspired by the analysis of distributional robustness, we have introduced a new approach where the weighting proportion is estimated based on the bound disagreement of each local model, rather than relying solely on sample proportions. To derive such bounds, we further utilize  the second-order moment over the robustness loss for more flat bound generalization, avoiding sharpness with nearly zero weighting on clients. This   
 has proven to be more effective in practical FL scenarios involving noise and class imbalance within FL. Extensive experimental results have further demonstrated that our novel weighting strategy significantly enhances the performance and robustness of typical FL algorithms. In future work, a self-adaptive weighting solution for gradient aggregation against the distributional training model is worth studying.


\appendices

\section{Proof of Bias-variance decomposition }
\label{appendix:A}

\textbf{Bias-variance decomposition.} Denoting $\mathbf{x}$ as the test sample and $y$ is the real label of $\mathbf{x}$, and dataset $D = \{(\mathbf{x}_1, y_1), \dots, (\mathbf{x}_n,y_n)\}$ is training set. The expected test error is decomposed as follows:
\begin{equation}
	\begin{aligned}
		& \underbrace{\mathbf{E}_{\mathbf{x}, y, D} \left[\left( y - h_{D}(\mathbf{x})  \right)^{2}\right]}_\mathrm{Expected\;Test\;Error} \\
		&=   \underbrace{\mathbf{E}_{\mathbf{x}, D}\left[\left(h_{D}(\mathbf{x}) - \bar{h}(\mathbf{x})\right)^{2}\right]}_\mathrm{Variance}\\
		&  + \underbrace{\mathbf{E}_{\mathbf{x}} \left[\left(\bar{h}(\mathbf{x}) - h(\mathbf{x})  \right)^{2}\right]}_\mathrm{Bias^2}  + \underbrace{\mathbf{E}_{\mathbf{x}, y} \left[\left( h(\mathbf{x}) -y\right) ^{2}\right]}_\mathrm{Noise}	.		
	\end{aligned}
\end{equation}
\textbf{Proof:}
\begin{equation}
	\begin{aligned}
		& \mathbf{E}_{\mathbf{x},y,D}\left[\left(y - h_{D}(\mathbf{x}) \right)^{2}\right] \\
		& = \mathbf{E}_{\mathbf{x},y,D}\left[\left[\left( y - \bar{h}(\mathbf{x})\right) + \left(\bar{h}(\mathbf{x}) - h_{D}(\mathbf{x})\right)\right]^{2}\right]  \\
		& = \mathbf{E}_{\mathbf{x}, D}\left[(y - \bar{h}(\mathbf{x}))^{2}\right]  + \mathbf{E}_{\mathbf{x}, y} \left[\left(\bar{h}(\mathbf{x}) - {h}_{D}(\mathbf{x})\right)^{2}\right] \\
		& + 2 \mathrm{\;} \mathbf{E}_{\mathbf{x}, y, D} \left[\left(\bar{h}(\mathbf{x})-h_{D}(\mathbf{x}) \right)\left( y-\bar{h}(\mathbf{x}) \right)\right],
	\end{aligned}
\end{equation}
The intermediate term can be simplified to 0:
\begin{equation*}
	\begin{aligned}
		& \mathbf{E}_{\mathbf{x}, y, D} \left[\left(\bar{h}(\mathbf{x})-h_{D}(\mathbf{x}) \right)\left( y-\bar{h}(\mathbf{x}) \right)\right]\\
		&= \mathbf{E}_{\mathbf{x}, y} \left[\mathbf{E}_{D} \left[ h_{D}(\mathbf{x}) - \bar{h}(\mathbf{x})\right] \left(\bar{h}(\mathbf{x}) - y \right) \right] \\
		&= \mathbf{E}_{\mathbf{x}, y} \left[ \left( \mathbf{E}_{D} \left[ h_{D}(\mathbf{x}) \right] - \bar{h}(\mathbf{x}) \right) \left(\bar{h}(\mathbf{x}) - y \right)\right] \\
		&= \mathbf{E}_{\mathbf{x}, y} \left[ \left(\bar{h}(\mathbf{x}) - \bar{h}(\mathbf{x}) \right) \left(\bar{h}(\mathbf{x}) - y \right)\right] \\
		&= \mathbf{E}_{\mathbf{x}, y} \left[ 0 \right] \\
		&= 0 ,
	\end{aligned}
\end{equation*}
Therefore, the initial formula can be simplified as:
\begin{equation}
	\begin{aligned}
		& \mathbf{E}_{\mathbf{x}, y, D} \left[ \left( y - h_{D}(\mathbf{x})  \right)^{2} \right] \\
		& = \underbrace{\mathbf{E}_{\mathbf{x}, D} \left[ \left( \bar{h}(\mathbf{x}) -  h_{D}(\mathbf{x}) \right)^{2} \right]}_\mathrm{Variance} + \mathbf{E}_{\mathbf{x}, y}\left[ \left( y - \bar{h}(\mathbf{x}) \right)^{2} \right] ,
	\end{aligned} 		 
\end{equation}      
Furthermore, the second term in the above equation is splited as follows:
\begin{equation}
	\begin{aligned}
		& \mathbf{E}_{\mathbf{x}, y} \left[ \left( \bar{h}(\mathbf{x}) -  y  \right)^{2}\right] \\
		&= \mathbf{E}_{\mathbf{x}, y} \left[ \left( (\bar{h}(\mathbf{x}) - h(\mathbf{x})) + (h(\mathbf{x}) - y) \right)^{2}\right]  \\
		&= \underbrace{\mathbf{E}_{\mathbf{x}} \left[\left(\bar{h}(\mathbf{x}) - h(\mathbf{x})  \right)^{2}\right]}_\mathrm{Bias^2} +\underbrace{\mathbf{E}_{\mathbf{x}, y} \left[\left( h(\mathbf{x}) -y\right) ^{2}\right]}_\mathrm{Noise}  \\  & + 2 \mathrm{\;} \mathbf{E}_{\mathbf{x}, y} \left[ (\bar{h}(\mathbf{x}) - h(\mathbf{x})) (h(\mathbf{x}) - y) \right] ,
	\end{aligned}
\end{equation}
The final term in above equation  is 0:
\begin{align*}
	& \mathbf{E}_{\mathbf{x}, y} \left[(\bar{h}(\mathbf{x}) - h(\mathbf{x})) (h(\mathbf{x}) - y)\right] \\
	&= \mathbf{E}_{\mathbf{x}}\left[\mathbf{E}_{y \mid \mathbf{x}} \left[ h(\mathbf{x}) - y \right] \left(\bar{h}(\mathbf{x}) - h(\mathbf{x}) \right) \right] \\
	&= \mathbf{E}_{\mathbf{x}} \left[ \left( h(\mathbf{x}) - \mathbf{E}_{y \mid \mathbf{x}} \left[ y \right]\right) \left(\bar{h}(\mathbf{x}) - h(\mathbf{x})\right)\right] \\
	&= \mathbf{E}_{\mathbf{x}} \left[ \left( h(\mathbf{x}) - h(\mathbf{x}) \right) \left(\bar{h}(\mathbf{x}) - h(\mathbf{x})\right)\right] \\
	&= \mathbf{E}_{\mathbf{x}} \left[ 0 \right] \\
	&= 0 ,
\end{align*}

Finally, the decomposition of expected test error as follows:
\begin{equation}
	\begin{aligned}
		& \underbrace{\mathbf{E}_{\mathbf{x}, y, D} \left[\left( y - h_{D}(\mathbf{x})  \right)^{2}\right]}_\mathrm{Expected\;Test\;Error} \\
		&= \underbrace{\mathbf{E}_{\mathbf{x}, D}\left[\left(h_{D}(\mathbf{x}) - \bar{h}(\mathbf{x})\right)^{2}\right]}_\mathrm{Variance}\\
		& + \underbrace{\mathbf{E}_{\mathbf{x}} \left[\left(\bar{h}(\mathbf{x}) - h(\mathbf{x})  \right)^{2}\right]}_\mathrm{Bias^2} + \underbrace{\mathbf{E}_{\mathbf{x}, y} \left[\left( h(\mathbf{x}) -y\right) ^{2}\right]}_\mathrm{Noise}	.		
	\end{aligned}
\end{equation}

\section{Proof of \textbf{Theorem 3.1} and \textbf{Theorem 3.2}}
\label{appendix:B}

\begin{theorem}[\textbf{Theorem3.1}]
	Upper bound of generalization performance for agnostic model under shifted distribution: Let $\mathcal{L}: Y \times Y \rightarrow \mathbf{R}_{+}$ be a loss function, for some $M>0$ assume that $\sup _{(x,y) \in (X,Y)}|\mathcal{L}(\mathbf{h}(x), y)| \leq M$ . Then $\sup _{(x,y) \in (X,Y)}|\mathcal{L}^2(\mathbf{h}(x), y)| \leq M^2$, for any probability measure $\mathbf{P}$ on (X,Y) and $\epsilon>0$, we have:
	\begin{equation*} 
		\begin{aligned}
			& \sup_{\mathbf{Q} \in B_\rho(\mathbf{P})} \mathbf{E}_\mathbf{Q}[\mathcal{L}^2(\mathbf{h}(X), Y)]  \leq  \mathbf{E}_\mathbf{P}[\mathcal{L}^2(\mathbf{h}(X), Y)]  + \\
			& 2 \lambda_\epsilon \left[\mathbf{V}_\mathbf{P}[\mathcal{L}^2(\mathbf{h}(X), Y)]\right]^{1/2} +\Delta^{u}_{\epsilon} ,
		\end{aligned}
	\end{equation*} 
	$$
	\begin{aligned}
		& \Delta^{u}_{\epsilon} = \epsilon^2\left(2-\epsilon^2\right)\left[ M^2-\mathbf{E}_\mathbf{P}[\mathcal{L}^2(\mathbf{h}(X), Y)]\right. \\ &\left.-\frac{\mathbf{V}_\mathbf{P}[\mathcal{L}^2(\mathbf{h}(X), Y)]}{M^2-\mathbf{E}_\mathbf{P}[\mathcal{L}^2(\mathbf{h}(X), Y)]}\right],	
	\end{aligned} 	
	$$
\end{theorem}
where $\lambda_\epsilon=\left[\epsilon^2\left(2-\epsilon^2\right)\left(1-\epsilon^2\right)^2\right]^{1/2}$ and $B_\epsilon(\mathbf{P})=\{\mathbf{Q} \in$$\mathbf{P}(X,Y): H(\mathbf{P}, \mathbf{Q}) \leq \epsilon\}$ is the Hellinger ball of radius $\epsilon$ centered at $\mathbf{P}$. The radius $\epsilon$ is required to be 
\begin{equation*}
	\epsilon^2 \leq 1-\left[ 1+ \left(\frac{M^2-\mathbf{E}_\mathbf{P}[\mathcal{L}^2(\mathbf{h}(X), Y)]}{\sqrt{\mathbf{V}_\mathbf{P}[\mathcal{L}^2(\mathbf{h}(X), Y)]}} \right)^2 \right]^{-1 / 2}.
\end{equation*}
\begin{theorem}[\textbf{Theorem3.2}]
	Lower bound of generalization performance for agnostic model under shifted distribution: Let $\mathcal{L}: Y \times Y \rightarrow \mathbf{R}_{+}$ be a nonnegative function taking values in $(X,Y)$. Then, for any probability measure $\mathbf{P}$ on $(X,Y)$ and $\epsilon>0$, we have:
	\begin{equation*}
		\begin{aligned}
			& \inf _{\mathbf{Q} \in B_\epsilon(\mathbf{P})} \mathbf{E}_\mathbf{Q}[\mathcal{L}^2(\mathbf{h}(X), Y)] \geq \mathbf{E}_\mathbf{P}[\mathcal{L}^2(\mathbf{h}(X), Y)] - \\
			& 2 \lambda_\epsilon \left[\mathbf{V}_\mathbf{P}[\mathcal{L}^2(\mathbf{h}(X), Y)]\right]^{1/2}-\Delta^{l}_{\epsilon} , 
		\end{aligned}		
	\end{equation*}
	\centering{$\Delta^{l}_{\epsilon} = \epsilon^2\left(2-\epsilon^2\right)\left[\mathbf{E}_\mathbf{P}[\mathcal{L}^2(\mathbf{h}(X), Y)]-\frac{\mathbf{V}_\mathbf{P}[\mathcal{L}^2(\mathbf{h}(X), Y)]}{\mathbf{E}_\mathbf{P}[\mathcal{L}^2(\mathbf{h}(X), Y)]}\right]$},
\end{theorem}
where $\lambda_\epsilon=\left[\epsilon^2\left(1-\epsilon^2\right)^2\left(2-\epsilon^2\right)\right]^{1/2}$ and $B_\epsilon(\mathbf{P})=\{\mathbf{Q} \in \mathcal{P}(X,Y): H(\mathbf{P}, \mathbf{Q}) \leq \epsilon\}$ is the Hellinger ball of radius $\epsilon$ centered at $\mathbf{Q}$. The radius $\epsilon$ is required to be small enough such that

\begin{equation*}
	\epsilon^2 \leq 1-\left[1+ \left( \frac{\mathbf{E}_\mathbf{P}[\mathcal{L}^2(\mathbf{h}(X), Y)]}{\sqrt{\mathbf{V}_\mathbf{P}[\mathcal{L}^2(\mathbf{h}(X), Y)]}} \right) ^2\right]^{-1 / 2} .
\end{equation*}
\textbf{Proof:} Starting from the determinant of the Gram matrix $G_f^2$~\cite{weinhold1968lower}:   
\begin{equation}
	\begin{aligned}
		& \left| G_{f^2} \right| := \left| \left[\begin{array}{c}\psi_Q \\ \psi_P \\ f^2 \psi_P\end{array}\right]\left[\begin{array}{lll}\psi_Q & \psi_P & f^2 \psi_P\end{array}\right] \right| \\\
		&= \left| \begin{array}{ccc}1 & \left\langle\psi_Q, \psi_P\right\rangle & \left\langle\psi_Q, f^2 \psi_P\right\rangle \\ \left\langle\psi_P, \psi_Q\right\rangle & 1 & \left\langle\psi_P, f^2 \psi_P\right\rangle \\ \left\langle f^2 \psi_P, \psi_Q\right\rangle  & \left\langle f^2 \psi_P, \psi_P\right\rangle & \left\langle f^2 \psi_P, f^2 \psi_P\right\rangle\end{array}\right| \\	
		& =  \left\langle f^2 \psi_P, f^2 \psi_P\right\rangle -\left\langle f^2 \psi_P, \psi_P\right\rangle^2 \\ 
		& +\left\langle f^2 \psi_P, \psi_Q\right\rangle\left(\left\langle\psi_P, \psi_Q\right\rangle\left\langle\psi_P, f^2 \psi_P\right\rangle-\left\langle f^2 \psi_P, \psi_Q\right\rangle\right)\\
		& -\left\langle\psi_Q, \psi_P\right\rangle \\
		& \times \left(\left\langle\psi_Q, \psi_P\right\rangle\left\langle f^2 \psi_P, f^2 \psi_P\right\rangle - \left\langle\psi_Q, f^2 \psi_P\right\rangle\left\langle f^2 \psi_P, \psi_P\right\rangle \right) \\
		& =  \left\langle f^2 \psi_P, f^2 \psi_P\right\rangle-\left\langle f^2 \psi_P, \psi_P\right\rangle^2 \\
		& +\left\langle f^2 \psi_P, \psi_Q\right\rangle\left\langle\psi_P, \psi_Q\right\rangle\left\langle\psi_P, f^2 \psi_P\right\rangle -\left\langle f^2 \psi_P, \psi_Q\right\rangle^2\\
		&-\left\langle\psi_Q, \psi_P\right\rangle^2\left\langle f^2 \psi_P, f^2 \psi_P\right\rangle \\
		& +\left\langle\psi_Q, \psi_P\right\rangle\left\langle\psi_Q, f^2 \psi_P\right\rangle\left\langle f^2 \psi_P, \psi_P\right\rangle \\
		& =  -\left\langle f^2 \psi_P, \psi_Q\right\rangle^2+2\left\langle\psi_P, \psi_Q\right\rangle\left\langle f^2 \psi_P, \psi_P\right\rangle\left\langle f^2 \psi_P, \psi_Q\right\rangle \\
		& +\left(1-\left\langle\psi_Q, \psi_P\right\rangle^2\right)\left\langle f^2 \psi_P, f^2 \psi_P\right\rangle -\left\langle f^2 \psi_P, \psi_P\right\rangle^2 ,
	\end{aligned}
\end{equation}     
For the sake of simplicity, let 
\\
$y=\left\langle f^2 \psi_P, \psi_Q\right\rangle$, $b_1=2\left\langle\psi_P, \psi_Q\right\rangle\left\langle f^2 \psi_P, \psi_P\right\rangle$,        
$c_1=\left(1-\left\langle\psi_Q, \psi_P\right\rangle^2\right)\left\langle f^2 \psi_P, f^2 \psi_P\right\rangle-\left\langle f^2 \psi_P, \psi_P\right\rangle^2$ , 
\\
We have
\begin{equation}
	G_f^2:=-x^2 + b_1x+ c_1,
\end{equation}
The property of gram matrix $\operatorname{det}(G_f^2) \geq 0$ means that
\begin{equation}
	\frac{b_1}{2}-\sqrt{\frac{b_1^2}{4}+c_1} \leq\left\langle f^2 \psi_P, \psi_Q\right\rangle \leq \frac{b_1}{2}+\sqrt{\frac{b_1^2}{4}+c_1},
\end{equation}
Where          
$$\frac{b_1}{2}=\left\langle\psi_P, \psi_Q\right\rangle\left\langle f^2\psi_P, \psi_P\right\rangle,$$
$$	
\begin{aligned} 
	& \sqrt{\frac{b_1^2}{4}+c_1}\\
	& =\left(\left\langle\psi_P, \psi_Q\right\rangle^2\left\langle f^2 \psi_P, \psi_P\right\rangle^2\right. \\
	& \left.+\left(1-\left\langle\psi_Q, \psi_P\right\rangle^2\right)\left\langle f^2 \psi_P, f^2 \psi_P\right\rangle-\left\langle f^2 \psi_P, \psi_P\right\rangle^2\right)^{\frac{1}{2}}   \\
	& =\left( \left\langle f^2 \psi_P, \psi_P\right\rangle^2\left(\left\langle\psi_P, \psi_Q\right\rangle^2-1\right)\right.\\
	& \left. +\left(1-\left\langle\psi_Q, \psi_P\right\rangle^2\right)\left\langle f^2 \psi_P, f^2 \psi_P\right\rangle \right)^{\frac{1}{2}} \\
	& =\sqrt{\left(1-\left\langle\psi_Q, \psi_P\right\rangle^2\right)\left(\left\langle f^2 \psi_P, f^2 \psi_P\right\rangle-\left\langle f^2 \psi_P, \psi_P\right\rangle^2\right)} ,\\
	&
\end{aligned}
$$
For simplity, we set \\
$\Delta F_1^2=\left\langle f^2 \psi_P, f^2 \psi_P\right\rangle-\left\langle f^2 \psi_P, \psi_P\right\rangle^2, \; S_1=\left\langle\psi_Q, \psi_P\right\rangle$, then $\sqrt{\frac{b^2}{4}+c}=\left(1-S_1^2\right)^{\frac{1}{2}}\left(\Delta F_1^2\right)^{\frac{1}{2}}$.
Thus, we get the following inequality
\begin{equation}
	\begin{aligned}
		& \left\langle f^2 \psi_P, \psi_Q\right\rangle \geq S_1\left\langle f^2 \psi_P, \psi_P\right\rangle-\left(1-S_1^2\right)^{\frac{1}{2}} \Delta F_1  \label{inequdelta} ,
	\end{aligned}
\end{equation}
According to Schwarz inequality
$$
\begin{aligned}
	& \left\langle f^2 \psi_P,\psi_Q\right\rangle \leq\left\langle\psi_Q, f^2 \psi_Q\right\rangle^{\frac{1}{2}}\left\langle\psi_P, f^2 \psi_P\right\rangle^{\frac{1}{2}} \\
	&\left\langle\psi_Q, f^2 \psi_Q\right\rangle
	\geq \frac{\left\langle f^2 \psi_P, \psi_Q\right\rangle^2}{\left\langle f^2 \psi_P, \psi_P\right\rangle} ,
\end{aligned}
$$
Finally, according to (\ref{inequdelta}), we get 
\begin{equation}
	\left\langle\psi_{Q,}, f^2 \psi_Q\right\rangle \geq \frac{\left\{S_1\left\langle f^2 \psi_P, \psi_P\right\rangle-\left(1-S_1^2\right)^{\frac{1}{2}} \Delta F_1\right\}^2}{\left\langle\psi_P, f^2 \psi_P\right\rangle} ,
\end{equation}
Furthermore, there are
\begin{equation}
	\begin{aligned}
		& \left\langle\psi_Q, f^2 \psi_Q\right\rangle \geq S_1^2\left\langle f^2 \psi_P, \psi_P\right\rangle \\
		& -2 S_1\left(1-S_1^2\right)^{\frac{1}{2}}\left\langle f^2 \psi_P, \psi_P\right\rangle \Delta F_1+\frac{\left(1-S_1^2\right) \Delta F_1^2}{\left\langle\psi_P, f^2 \psi_P\right\rangle}, \label{inequF}
	\end{aligned}
\end{equation}       
According to the definition	                      
$$
\begin{aligned}
	\psi_P:=\sqrt{\frac{d P}{d \mu}}, \; \psi_Q:=\sqrt{\frac{d Q}{d \mu}}\;, \langle f, g\rangle_{L_2}:=\int_z f g d \mu ,\\ \quad M_{f{^2}} \psi: Z \mapsto\left(M_{f{^2}} \psi\right)(z)=f^2(z) \cdot \psi(z),
\end{aligned}
$$	 
We have 
$$	
\begin{aligned}
	\mathbf{E}_{Z \sim P}[f^2(z)]=\int_Z f^2(z) d p(z)=\int_Z f^2(z) \frac{d p}{d \mu}(z) d \mu(z) \\
	=\int_Z \psi_P(z) f^2(z) \psi_P(z) d \mu(z) =\left\langle\psi_P, M_{f{^2}} \psi_P\right\rangle_{L_2},
\end{aligned}						 
$$
Similarly, we get the variance of $f(Z)^2$
$$	 						 
\mathbf{V}_{Z \sim p}[f^2(Z)]=\left\langle\psi_P, M_{f^2} \psi_P\right\rangle_{L_2}-\left\langle\psi_P, M_{f^2} \psi_P\right\rangle_{L_2}^2 .
$$
\subsection{Upper bound} 

\textbf{Upper bound of $\mathbf{E}_{Q}[l^2(Z)]$}. Given $\sup_{z\in Z}|l^2(z)|\leq M^2$, we can observe that the function $f^2_{l}(\cdot) = M^2 - l^2(\cdot)$ is bounded and, consequently, defines a bounded linear operator. By substituting the items in inequality (\ref{inequF}), we can establish the resulting bound as follows:

$$
\begin{aligned}
	&\mathbf{E}_Q[M^2-l^2(Z)] \geq\left|\left\langle\psi_P, \psi_Q\right\rangle\right|^2 \mathbf{E}_P[M^2-l^2(Z)]\\
	& -2\left|\left\langle\psi_P, \psi_Q\right\rangle\right| \sqrt{\left.\left(1-\left|\left\langle\psi_P, \psi_Q\right\rangle\right|^2\right) \mathbf{V}_P[M^2-l^2(Z))\right]} \\
	&+\frac{\left(1-\left|\left\langle\psi_P, \psi_P\right\rangle\right|^2\right) \mathbf{V}_P[M^2-l^2(Z)]}{\mathbf{E}_P[M^2-l^2(Z)]},
\end{aligned}	
$$
Furthermore, there are
\begin{equation}
	\begin{aligned}
		&\mathbf{E}_Q[l^2(Z)] \leq\left|\left\langle\psi_P, \psi_Q\right\rangle\right|^2 \mathbf{E}_P[l^2(Z)-M^2] + M^2 \\ &+2\left|\left\langle\psi_P, \psi_Q\right\rangle\right| \sqrt{\left.\left(1-\left|\left\langle\psi_P, \psi_Q\right\rangle\right|^2\right) \mathbf{V}_P[M^2-l^2(Z)\right]} \\
		&-\frac{\left(1-\left|\left\langle\psi_P, \psi_P\right\rangle\right|^2\right) \mathbf{V}_P[M^2-l^2(Z)]}{\mathbf{E}_P[M^2-l^2(Z)]}, \label{inequCo}
	\end{aligned}	
\end{equation}
To simplify Eq (\ref{inequCo}), we separately calculate some of its terms
\begin{equation*}
	\begin{aligned}	
		\mathbf{V}_P[M^2-l^2(Z)]=\mathbf{V}_P[l^2(Z)], \; \\ \left\langle\psi_P, \psi_Q\right\rangle=1-H^2(P, Q)=1-\rho^2,
	\end{aligned}
\end{equation*}
\begin{equation*}
	\begin{aligned}	
		& H^2(P, Q)=\frac{1}{2} \int_Z\left(\psi_P-\psi_Q\right)^2 d \mu \\
		& =1-\int_Z \psi_P \psi_Q d \mu=1-\left\langle\psi_P, \psi_Q\right\rangle,\\
		& \left(1-\left(1-\rho^2\right)^2\right) \mathbf{E}_P\left[l^2(Z)-M^2\right] \\
		& =\rho^2\left(2-\rho^2\right) \mathbf{E}_P\left[l^2(Z)-M^2\right] ,\\     
		& \left(\left(1-\rho^2\right)^2-1\right) \mathbf{E}_P[l^2(Z)-M^2]\\
		& =\left(1-\rho^2\right)^2 \mathbf{E}_P\left[(l^2(Z)-M^2]-\mathbf{E}_P[(l^2(Z)-M^2]\right. \\    
		& 2\left(1-\rho^2\right) \sqrt{\left(1-\left(1-\rho^2\right)^2\right) \mathbf{V}_P[l^2(Z)]}\\
		& =2 \sqrt{\rho^2\left(2-\rho^2\right)\left(1-\rho^2\right)^2 \mathbf{V}_P[l^2(Z)]}  ,      	
	\end{aligned}
\end{equation*}    
Substitute the above equation into Eq (\ref{inequCo}), we have
$$
\begin{aligned}
	& \mathbf{E}_Q[l^2(Z)] \leq\left(1-\rho^2\right)^2 \mathbf{E}_P[l^2(Z)-M^2] + M^2 \\
	&+ 2\left(1-\rho^2\right) \sqrt{\left(1-\left(1-\rho^2\right)^2\right) \mathbf{V}_P[l^2(Z)]} \\ 
	&-\frac{\left(1-\left(1-\rho^2\right)^2\right) \mathbf{V}_P[l^2(Z)]}{\mathbf{E}_P[M^2-l^2(Z)]} \\
	& =\mathbf{E}_P[l^2(Z)-M^2]+ (1-\rho^2)^2 \mathbf{E}_P[l^2(Z)-M^2]\\
	& -\mathbf{E}_P[l^2(Z)-M^2]+ M^2 +2 C_{\rho} \sqrt{\mathbf{V}_P[l^2(Z)]}\\
	&-\frac{\rho^2\left(2-\rho^2\right) \mathbf{V}_P[l^2(Z)]}{\mathbf{E}_P[M^2-l^2(Z)]} \\
	& =\mathbf{E}_P[l^2(Z)]+2 C_{\rho} \sqrt{\mathbf{V}_P[l^2(Z)]}\\
	& + \rho^2\left(2-\rho^2\right) \left[ M^2-\mathbf{E}_P[l^2(Z)]-\frac{ \mathbf{V}_P[l^2(Z)]}{\mathbf{E}_P[M^2-l^2(Z)]} \right] ,
\end{aligned}
$$
Where $C_\rho=\sqrt{\rho^2\left(2-\rho^2\right)\left(1-\rho^2\right)^2}$.
Finally, we get the upper bound of $\mathbf{E}_Q[l^2(Z)]$:
\begin{equation}
	\begin{aligned}
		& \mathbf{E}_Q[l^2(Z)] \leq \mathbf{E}_P[l^2(Z)]+2 C \rho \sqrt{\mathbf{V}_P[l^2(Z)]} \\
		& +\rho^2\left(2-\rho^2\right)\left[M^2-\mathbf{E}_P[l^2(Z)]-\frac{\mathbf{V}_P[l^2(Z)]}{\mathbf{E}_P[M^2-l^2(Z)]}\right] ,
	\end{aligned}
	\label{inequfin}
\end{equation}    
From right side of inequality (\ref{inequdelta}), we have
$$
\frac{S_1}{\left(1-S_1^2\right)^{\frac{1}{2}}} \geq \frac{\Delta F_1}{\left\langle f^2 \psi_P, \psi_P\right\rangle},
$$
It means that
$$
\frac{\left|\left\langle\psi_P, \psi_Q\right\rangle\right|^2}{1-\left|\left\langle\psi_P, \psi_Q\right\rangle\right|^2} \geq \frac{\mathbf{V}_P[l^2(Z)]}{\mathbf{E}_P^2[M^2-l^2(Z)]} ,
$$
Furthermore,
$$
\begin{aligned}        	
	& \sqrt{\frac{\left|\left\langle\psi_P, \psi_Q\right\rangle\right|^2}{1-\left|\left\langle\psi_P, \psi_Q\right\rangle\right|^2}} \geq \frac{\sqrt{\mathbf{V}_P[l^2(Z)]}}{\mathbf{E}_P[M^2-l^2(Z)]} \\ 
	& \Longrightarrow \sqrt{\frac{1-\left|\left\langle\psi_P, \psi_Q\right\rangle\right|^2}{\left|\left\langle\psi_P, \psi_Q\right\rangle\right|^2}} \leq \frac{M^2-\mathbf{E}_P[l^2(Z)]}{\sqrt{\mathbf{V}_P[l^2(Z)]}} \\
	& \Longrightarrow  \frac{1}{\left\langle\psi_P, \psi_Q\right\rangle^2} \leq 1+\left(\frac{M^2-\mathbf{E}_P[l^2(Z)]}{\sqrt{\mathbf{V}_P[l^2(Z)]}}\right)^2 \\
	& \xrightarrow[]{\left\langle\psi_P, \psi_Q\right\rangle=1-\rho^2} \frac{1}{\left(1-\rho^2\right)^2} \leq 1+\left(\frac{M^2-\mathbf{E}_P[l^2(Z)]}{\sqrt{\mathbf{V}_P[l^2(Z)]}}\right)^2  \\
	&  \Longrightarrow 1-\rho^2 \geq\left[1+\left(\frac{M^2-\mathbf{E}_P[l^2(Z)]}{\sqrt{\mathbf{V}_P[l^2(Z)]}}\right)^2\right]^{-\frac{1}{2}} \\
	&  \Longrightarrow \rho^2 \leq 1-\left[1+\left(\frac{M^2-\mathbf{E}_P[l^2(Z)]}{\sqrt{\mathbf{V}_P[l^2(Z)]}}\right)^2\right]^{-\frac{1}{2}}, \\
\end{aligned}
$$
The conditions that need to be met for the above eq (\ref{inequfin})
\begin{equation}
	\rho^2 \leq 1-\left[1+\left(\frac{M^2-\mathbf{E}_P[l^2(Z)]}{\sqrt{\mathbf{V}_P[l^2(Z)]}}\right)^2\right]^{-\frac{1}{2}}.
\end{equation}
\subsection{Lower bound}        
\textbf{Lower bound of $\mathbf{E}_{Q}[l^2(Z)]$}.
Given $\sup_{z\in Z}|l(z)|\leq M$, we can conclude that the function $f_{l}(\cdot) = l(\cdot)$ is bounded and therefore defines a bounded linear operator. By substituting the items in inequality (\ref{inequF}), we obtain the lower bound as follows:

\begin{equation}
	\begin{aligned}
		&\mathbf{E}_Q[l^2(Z)] \geq\left|\left\langle\psi_P, \psi_Q\right\rangle\right|^2 \mathbf{E}_P[l^2(Z)]\\
		& -2\left|\left\langle\psi_P, \psi_Q\right\rangle\right| \sqrt{\left(1-\left|\left\langle\psi_P, \psi_Q\right\rangle\right|^2\right) \mathbf{V}_P[l^2(Z)]} \\
		&+\frac{\left(1-\left|\left\langle\psi_P, \psi_P\right\rangle\right|^2\right) \mathbf{V}_P[l^2(Z)]}{\mathbf{E}_P[l^2(Z)]} ,  \label{equlower}
	\end{aligned}	
\end{equation}

Same as the upper bound, we separately calculate some of its terms
$$
\begin{aligned}	
	\left\langle\psi_P, \psi_Q\right\rangle=1-H^2(P, Q)=1-\rho^2,
\end{aligned}
$$
$$
\begin{aligned}	
	& H^2(P, Q)=\frac{1}{2} \int_Z\left(\psi_P-\psi_Q\right)^2 d \mu=1-\int_Z \psi_P \psi_Q d \mu \\
	& =1-\left\langle\psi_P, \psi_Q\right\rangle,
\end{aligned}
$$       
$$
\begin{aligned}	
	{\left(1-\left(1-\rho^2\right)^2\right) \mathbf{E}_P\left[l^2(Z)]=\rho^2\left(2-\rho^2\right) \mathbf{E}_P[l^2(Z)]\right.}, 
\end{aligned}
$$
$$
\begin{aligned}	
	& \left(\left(1-\rho^2\right)^2-1\right) \mathbf{E}_P[l^2(Z)]\\
	& =\left(1-\rho^2\right)^2 \mathbf{E}_P\left[(l^2(Z)]-\mathbf{E}_P[l^2(Z)]\right. ,
\end{aligned}
$$
$$
\begin{aligned}	
	& 2\left(1-\rho^2\right) \sqrt{\left(1-\left(1-\rho^2\right)^2\right) \mathbf{V}_P[l^2(Z)]}\\
	& =2 \sqrt{\rho^2\left(2-\rho^2\right)\left(1-\rho^2\right)^2 \mathbf{V}_P[l^2(Z)]} ,   	
\end{aligned}
$$        
Substitute the items in inequality (\ref{equlower}), we have:
$$
\begin{aligned}
	& \mathbf{E}_Q[l^2(Z)] \geq\left(1-\rho^2\right)^2 \mathbf{E}_P[l^2(Z)] \\
	& - 2\left(1-\rho^2\right) \sqrt{\left(1-\left(1-\rho^2\right)^2\right) \mathbf{V}_P[l^2(Z)]} \\
	& +\frac{\left(1-\left(1-\rho^2\right)^2\right) \mathbf{V}_P[l^2(Z)]}{\mathbf{E}_P[l^2(Z)]} \\
	& =\mathbf{E}_P[l^2(Z)]+ (1-\rho^2)^2 \mathbf{E}_P[l^2(Z)]-\mathbf{E}_P[l^2(Z)]\\
	& -2 C_{\rho} \sqrt{\mathbf{V}_P[l^2(Z)]} -\frac{\left.\rho^2\left(2-\rho^2\right) \mathbf{V}_P[l^2(Z)]\right]}{\mathbf{E}_P[l^2(Z)]} \\
	& =\mathbf{E}_P[l^2(Z)]-2 C_{\rho} \sqrt{\mathbf{V}_P[l^2(Z)]}- \\
	& \rho^2\left(2-\rho^2\right) \left[\mathbf{E}_P[l^2(Z)]-\frac{ \mathbf{V}_P[l^2(Z)]}{\mathbf{E}_P[l^2(Z)]} \right] ,
\end{aligned}
$$
where $C_\rho=\sqrt{\rho^2\left(2-\rho^2\right)\left(1-\rho^2\right)^2}$.  Finally, we get the lower bound of $\mathbf{E}_Q[l^2(Z)]$:
\begin{equation}
	\begin{aligned}
		& \mathbf{E}_Q[l^2(Z)] \geq \mathbf{E}_P[l^2(Z)]-2 C \rho \sqrt{\mathbf{V}_P[l^2(Z)]} \\
		& -\rho^2\left(2-\rho^2\right)\left[\mathbf{E}_P[l^2(Z)]-\frac{\mathbf{V}_P[l^2(Z)]}{\mathbf{E}_P[l^2(Z)]}\right] ,
	\end{aligned}
	\label{ineqLF}
\end{equation}
From right side of inequality (\ref{inequdelta}), we have
\begin{equation*}
	\frac{S}{\left(1-S^2\right)^{\frac{1}{2}}} \geq \frac{\Delta F}{\left\langle f \psi_P, \psi_P\right\rangle} ,
\end{equation*}
It means that
\begin{equation*}
	\frac{\left|\left\langle\psi_P, \psi_Q\right\rangle\right|^2}{1-\left|\left\langle\psi_P, \psi_Q\right\rangle\right|^2} \geq \frac{\mathbf{V}_P[l^2(Z)]}{\mathbf{E}_P^2[l^2(Z)]} ,
\end{equation*}
Furthermore
$$
\begin{aligned}        	
	& \sqrt{\frac{\left|\left\langle\psi_P, \psi_Q\right\rangle\right|^2}{1-\left|\left\langle\psi_P, \psi_Q\right\rangle\right|^2}}Z \geq \frac{\sqrt{\mathbf{V}_P[l^2(Z)]}}{\mathbf{E}_P[l^2(Z)]} \\
	& \Longrightarrow \sqrt{\frac{1-\left|\left\langle\psi_P, \psi_Q\right\rangle\right|^2}{\left|\left\langle\psi_P, \psi_Q\right\rangle\right|^2}} \leq \frac{\mathbf{E}_P[l^2(Z)]}{\sqrt{\mathbf{V}_P[l^2(Z)]}} \\
	& \Longrightarrow \frac{1}{\left\langle\psi_P, \psi_Q\right\rangle^2} \leq 1+\left(\frac{\mathbf{E}_P[l^2(Z)]}{\sqrt{\mathbf{V}_P[l^2(Z)]}}\right)^2 \\
	& \xrightarrow[]{\left\langle\psi_P, \psi_Q\right\rangle=1-\rho^2} \frac{1}{\left(1-\rho^2\right)^2} \leq 1+\left(\frac{\mathbf{E}_P[l^2(Z)]}{\sqrt{\mathbf{V}_P[l^2(Z)]}}\right)^2 \\
	& \Longrightarrow  1-\rho^2 \geq \left[1+\left(\frac{\mathbf{E}_P[l^2(Z)]}{\sqrt{\mathbf{V}_P[l^2(Z)]}}\right)^2\right]^{-\frac{1}{2}}\\ & \Longrightarrow  \rho^2 \leq 1-\left[1+\left(\frac{\mathbf{E}_P[l^2(Z)]}{\sqrt{\mathbf{V}_P[l^2(Z)]}}\right)^2\right]^{-\frac{1}{2}} , \\
\end{aligned}
$$
The conditions that need to be met for the above Eq (\ref{ineqLF})
\begin{equation}
	\rho^2 \leq 1-\left[1+\left(\frac{\mathbf{E}_P[(l^2(Z)]}{\sqrt{\mathbf{V}_P[l^2(Z)]}}\right)^2\right]^{-\frac{1}{2}} .
\end{equation}

\section{The estimation of $\mathbf{E}_\mathbf{P}$,$\mathbf{V}_\mathbf{P}$ }
\label{appendix:C}

The estimation of $\mathbf{E}_\mathbf{P}[\mathcal{L}^2(\mathbf{h}(X), Y)]$ and $\mathbf{V}_\mathbf{P}[\mathcal{L}^2(\mathbf{h}(X), Y)]$ is key in \textbf{Theorem 3.1} and \textbf{Theorem 3.2}. The estimation methods mentioned in Werber et al.~\cite{weber2022certifying} are shown as below.
\begin{corollary}
	(Hoeffding,1963)~\cite{hoeffding1963probability}  Let $(X_1,Y_1)$, $\ldots,$ $(X_n,Y_n)$ be independent random variables drawn from $\mathbf{P}$ and taking values in $(X,Y)$. Let $\mathcal{L}: (\mathbf{h}(X),Y) \rightarrow[0, M]$ be a loss function and let $\hat{\mathcal{L}^2}_n:=\frac{1}{n} \sum_{i=1}^n \mathcal{L}^2\left(\mathbf{h}(X_i),Y_i\right)$ be the mean under the empirical distribution $\hat{\mathbf{P}}_n$. Then for $\delta>0$, with probability at least $1-\delta$, 
	\begin{equation}
		\mathbf{E}_\mathbf{P}[\mathcal{L}^2(\mathbf{h}(X), Y)] \leq \hat{\mathcal{L}^2}_n +M \left[\frac{\ln 1 / \delta}{2 n}\right]^{1/2} ,
	\end{equation}
\end{corollary}

\begin{corollary}
	(Maurer \& Pontil,2009)~\cite{maurer2009empirical}  Let $(X_1,Y_1)$, $\ldots$, $(X_n,Y_n)$ be independent random variables drawn from distribution $\mathbf{P}$ and taking values in $(X,Y)$. For a loss function $\mathcal{L}: (\mathbf{h}(X_i),Y) \rightarrow[0, M]$, let 
	\begin{equation*}
		\begin{aligned}
			S_n^2:=\frac{1}{n(n-1)} \sum_{1 \leq i<j \leq n}^n\left(\mathcal{L}^2\left(\mathbf{h}(X_i),Y_i\right)-\mathcal{L}^2\left(\mathbf{h}(X_j),Y_j\right)\right)^2 ,
		\end{aligned}
	\end{equation*}
	be the unbiased estimator of the variance of the random variable $\mathcal{L}(\mathbf{h}(X),Y)$, $(X,Y) \sim \mathbf{P}$. Then for $\delta>0$, with probability at least $1-\delta$,
	\begin{equation}
		\left[\mathbf{V}_\mathbf{P}[\mathcal{L}^2(\mathbf{h}(X), Y)]\right]^{1/2} \leq \left[S_n^2\right]^{1/2}+M^2 \left[\frac{2 \ln 1 / \delta}{n-1}\right]^{1/2} .
	\end{equation}
\end{corollary}

\bibliographystyle{IEEEtran}


\end{document}